\documentclass[letterpaper]{article} 
\usepackage{aaai24}  
\usepackage{times}  
\usepackage{helvet}  
\usepackage{courier}  
\usepackage[hyphens]{url}  
\usepackage{graphicx} 
\usepackage{natbib}  
\usepackage{caption} 
\usepackage[utf8]{inputenc} 
\usepackage[T1]{fontenc}    
\usepackage{url}            
\usepackage{booktabs}       
\usepackage{amsfonts}       
\usepackage{nicefrac}       
\usepackage{microtype}      
\usepackage{xcolor}         

\usepackage[noend,ruled,linesnumbered]{algorithm2e}
\DontPrintSemicolon 
\SetKwInput{KwInput}{Input}
\SetKwInput{KwOutput}{Output}
\usepackage{subcaption}
\usepackage{enumitem}
\usepackage{mathtools}
\usepackage{tikz}
\usepackage{amsmath}
\usepackage{amssymb}
\usepackage{amsthm}
\usetikzlibrary{positioning, fit, backgrounds, calc, decorations, decorations.markings, shapes, shapes.misc, arrows.meta, shadows}
\usepackage{multirow}
\usepackage{float}
\usepackage{standalone}
\usepackage{import}
\usepackage{bbold}

\usepackage{pgfplots}
\pgfplotsset{compat=newest}

\usepackage{pifont}

\def\myparagraph#1{\vspace*{3pt}\noindent{\bf #1~~}}

\DeclareMathOperator*{\R}{\mathbb{R}}

\DeclarePairedDelimiter\abs{\lvert}{\rvert}%
\DeclareMathOperator{\argmin}{argmin}

\DeclareMathOperator{\shp}{SP}

\providecommand{\la}{\langle}
\providecommand{\ra}{\rangle}

\providecommand{\abs}[1]{\lvert #1 \rvert}

\renewcommand{\P}{\mathcal{P}}
\providecommand{\M}{\mathcal{M}}
\providecommand{\X}{\mathcal{X}}
\providecommand{\I}{\mathcal{I}}
\providecommand{\J}{\mathcal{J}}
\providecommand{\E}{\mathcal{E}}
\providecommand{\old}{\textnormal{\texttt{in}}}
\providecommand{\new}{\textnormal{\texttt{out}}}
\providecommand{\LN}{\textnormal{\texttt{LN}}}
\providecommand{\MP}{\textnormal{\texttt{CONV}}}

\providecommand{\der}[1]{\dot{#1}}

\newcommand{\0}{\mathbb{0}}

\definecolor{varcolor}{HTML}{000000}
\definecolor{concolor}{HTML}{000000}
\definecolor{edgecolor}{HTML}{000000}
\definecolor{bgcolor}{HTML}{E2E2E2}
\definecolor{arrowcolor}{HTML}{474747}
\definecolor{gnncolor}{HTML}{DAA9D0}
\definecolor{dmmacolor}{HTML}{ECDCB0}
\definecolor{dmmacolor2}{HTML}{EDC966}
\definecolor{nonparamcolor}{HTML}{C1D7AE}
\definecolor{losscolor}{HTML}{A3CBFF}

\definecolor{customcolor1}{HTML}{7570b3}
\definecolor{customcolor2}{HTML}{FF7B9C}
\definecolor{customcolor3}{HTML}{d95f02}
\definecolor{customcolor4}{HTML}{1b9e77}
\definecolor{customcolor5}{HTML}{AA4499}
\pgfplotsset{
dashstar/.style={dash pattern=on 8pt off 8pt,
postaction={decorate,decoration={
   markings,
   mark=between positions 12pt and 1 step 16pt with {
     \node[line width = 0.75pt] {\pgfuseplotmark{star}};
   }}}},
dashcircle/.style={dash pattern=on 8pt off 8pt,
postaction={decorate,decoration={
   markings,
   mark=between positions 12pt and 1 step 16pt with {
     \node[line width = 0.75pt] {\pgfuseplotmark{*}};
   }}}},
dashsquare/.style={dash pattern=on 8pt off 8pt,
postaction={decorate,decoration={
   markings,
   mark=between positions 12pt and 1 step 16pt with {
     \node[line width = 0.75pt] {\pgfuseplotmark{square}};
   }}}},
dashdiamond/.style={dash pattern=on 8pt off 8pt,
postaction={decorate,decoration={
   markings,
   mark=between positions 12pt and 1 step 16pt with {
     \node[line width = 0.75pt] {\pgfuseplotmark{triangle}};
   }}}},
dashtri/.style={dash pattern=on 7pt off 9pt,
postaction={decorate,
decoration={markings,
  mark=between positions 12pt and 1 step 16pt with {
     \node[line width = 0.75pt] {\pgfuseplotmark{diamond*}};
  }}}}}
  
\usepackage{pgfplotstable}
\usepackage{longtable}
\pgfplotstableset{col sep=comma}
\pgfkeys{/pgf/number format/.cd, set thousands separator={}}
\pgfplotstableset{empty header/.style={every head row/.style={output empty row},}}
\newcommand{\insertTable}[3]{
\pgfplotstabletypeset[
    empty header,
    begin table=\begin{longtable},
    every first row/.append style={before row={%
    \caption{#2}%
    \label{#3}\\ 
    \toprule
    instance & method & \multicolumn{4}{c}{Until termination criteria} & \multicolumn{4}{c}{Best until max.\ num itr.} \\
    \cmidrule(lr){3-6} \cmidrule(lr){7-10}
    &  & 
    $E$ ($\uparrow$) & $g(t)$ ($\downarrow$) & $t$ ($\downarrow$) & \# itr. &
    $E$ ($\uparrow$) & $g(t)$ ($\downarrow$) & $t$ ($\downarrow$) & \# itr.\\ \toprule    
    \endfirsthead
    \multicolumn{10}{c}%
    {{\bfseries Table \thetable\ Continued from previous page}} \\
    \toprule
    instance & method & \multicolumn{4}{c}{Until termination criteria} & \multicolumn{4}{c}{Best until max.\ num itr.} \\
    \cmidrule(lr){3-6} \cmidrule(lr){7-10}
    &  & 
    $E$ ($\uparrow$) & $g(t)$ ($\downarrow$) & $t$ ($\downarrow$) & \# itr. &
    $E$ ($\uparrow$) & $g(t)$ ($\downarrow$) & $t$ ($\downarrow$) & \# itr.\\ \toprule
    \endhead
    \midrule \multicolumn{10}{r}{\textit{Continued on next page}} \\ \bottomrule
    \endfoot
    \bottomrule
    \endlastfoot
    }},
    every nth row={3}{before row=\midrule},
    end table=\end{longtable},
    col sep=comma,
	columns/instance_name/.style={string type, column type=l, 
	string replace*={_}{-}},
	columns/method/.style={string type, column type=l,
    	string replace*={Gurobi}{\texttt{Gurobi}}, string replace*={FastDOG}{\texttt{FastDOG}},
    	string replace*={DOGEM}{\normalfont{\texttt{DOGE-M}}},
    	string replace*={DOGE}{\texttt{DOGE}}, 
	},
	string replace*={_}{-},
 	columns/lb_es/.style={fixed, precision=0, empty cells with={-}},
 	columns/lb_last/.style={fixed, precision=0, empty cells with={-}},
 	columns/gap_es/.style={fixed, precision=5, empty cells with={-}},
 	columns/gap_last/.style={fixed, precision=5, empty cells with={-}},
  	columns/time_es/.style={fixed, precision=1,empty cells with={-}},
  	columns/time_last/.style={fixed, precision=1, empty cells with={-}},
  	columns/itr_es/.style={fixed, precision=1,empty cells with={-}},
  	columns/itr_last/.style={fixed, precision=1,empty cells with={-}},
	column type = r,
]{#1}
}
\usepackage{collcell}
\usepackage{diagbox}

\newtoggle{inTableHeader}
\toggletrue{inTableHeader}
\newcommand*{\StartTableHeader}{\global\toggletrue{inTableHeader}}%
\newcommand*{\EndTableHeader}{\global\togglefalse{inTableHeader}}%

\let\OldTabular\tabular%
\let\OldEndTabular\endtabular%
\renewenvironment{tabular}{\StartTableHeader\OldTabular}{\OldEndTabular\StartTableHeader}%
\newcommand*{\MinNumber}{0.0}%
\newcommand*{\MidNumber}{0.01} %
\newcommand*{\MaxNumber}{1.0}%

\newcommand{\ApplyGradient}[1]{%
  \iftoggle{inTableHeader}{#1}{
    \ifdim #1 pt > \MidNumber pt
        \pgfmathsetmacro{\PercentColor}{max(min(100.0*(#1 - \MidNumber)/(\MaxNumber-\MidNumber),100.0),0.00)} %
        \hspace{-0.33em}\colorbox{red!\PercentColor!yellow}{#1}
    \else
        \pgfmathsetmacro{\PercentColor}{max(min(100.0*(\MidNumber - #1)/(\MidNumber-\MinNumber),100.0),0.00)} %
        \hspace{-0.33em}\colorbox{green!\PercentColor!yellow}{#1}
    \fi
  }}
\newcolumntype{R}{>{\collectcell\ApplyGradient}c<{\endcollectcell}}

\theoremstyle{plain}

\newtheorem{prop}{Proposition}

\theoremstyle{definition}
\newtheorem{defn}{Definition}

\theoremstyle{remark}
\newtheorem*{rem}{Remark}

\title{DOGE-Train: Discrete Optimization on GPU with End-to-end Training}
\author {
Ahmed Abbas\textsuperscript{\rm 1} \hspace{5cm} Paul Swoboda\textsuperscript{\rm 1,\rm 2,\rm 3}
}
\affiliations {
 \textsuperscript{\rm 1}Max Planck Institute for Informatics, Saarland Informatics Campus\\
  \textsuperscript{\rm 2}University of Mannheim, 
  \textsuperscript{\rm 3}Heinrich-Heine University D\"usseldorf
}

\begin{document}

\maketitle

\begin{abstract}
We present a fast, scalable, data-driven approach for solving relaxations of 0-1 integer linear programs.
We use a combination of graph neural networks (GNN) and the Lagrange decomposition based algorithm~\cite{abbas2021fastdog}.
We make the latter differentiable for end-to-end training and use GNNs to predict its algorithmic parameters.
This allows to retain the algorithm's theoretical properties including dual feasibility and guaranteed non-decrease in the lower bound while improving it via training.
We overcome suboptimal fixed points of the basic solver by additional non-parametric GNN update steps maintaining dual feasibility.
For training we use an unsupervised loss.
We train on smaller problems and test on larger ones showing strong generalization performance with a GNN comprising only around $10k$ parameters.
Our solver achieves significantly faster performance and better dual objectives than its non-learned version, achieving close to optimal objective values of LP relaxations of very large structured prediction problems and on selected combinatorial ones.
In particular, we achieve better objective values than specialized approximate solvers for specific problem classes while retaining their efficiency.
Our solver has better any-time performance over a large time period compared to  a commercial solver. 
\end{abstract}

\section{Introduction}
\label{sec:introduction}

Integer linear programs (ILP) are a universal tool for solving combinatorial optimization problems.
While great progress has been made on improving ILP solvers over the past several decades, there is recent interest in leveraging machine learning to enhance ILP algorithms.
Almost all ILP solving subroutines, except ILP relaxation algorithms, have been recently shown to benefit from learning, including variable selection for branch-and bound~\cite{nair2020solving} or cutting plane selection~\cite{huang2022learning,turner2022adaptive,paulus2022learning_cutting}.
Moreover, a number of specialized heuristics as well as meta-algorithms using heuristics as subroutines~\cite{sun2023difusco,qiu2022dimes} have used ML for greatly improving performance for some problem classes.
However no general purpose ILP relaxation algorithm has yet benefited from machine learning. 

We make a contribution towards general ML-enabled solvers for optimization by learning a problem-agnostic solver for LP-relaxations of ILPs.
LP solving is a key step taking most time in traditional ILP pipelines.
State of the art LP solvers~\cite{gurobi,cplex,fico,mosek,SCIP} are not amenable to ML since they are non-differentiable, sequential and have very complex implementations.
This makes utilization of neural networks and GPUs for solver improvement difficult.
For these reasons we build upon the massively parallel FastDOG~\cite{abbas2021fastdog} solver and show that it can be made differentiable. 
This allows to train our problem agnostic solver for specific problem classes resulting in equal or better performance as compared to efficient hand-designed specialized solvers.

\paragraph{Contributions}
Our high-level contributions are 
conceptual and empirical:
(i)~We show that embedding good inductive biases coming from non-learned solvers (in our case highly parallel GPU-based block coordinate ascent~\cite{abbas2021fastdog} and subgradients) into neural networks leads to greatly improved performance. 
In particular, we give evidence to the hypothesis that similar to vision (convolutions) and NLP (sequence models) the right inductive biases coming from solver primitives are a promising way to use the potential of ML for optimization. 
(ii)~Our approach is more economical as compared to developing efficient problem specific heuristics, as is customary for large scale problems in structured prediction tasks for ML~\cite{haller2020primal,hutschenreiter2021fusion_gm}. Instead of spending much time and effort in designing and implementing new algorithms, one can train our problem agnostic solver with a few problem instances coming from the problem class of interest and obtain a state of the art GPU-enabled solver for it\footnote{Code available at \url{https://github.com/LPMP/BDD}}.

In detail, we propose to learn the Lagrange decomposition algorithm~\cite{abbas2021fastdog} for solving LP relaxations of ILP problems and show its benefits.
In particular,
\begin{itemize}[itemsep=1pt,parsep=1pt,leftmargin=0.25cm]
\item 
We generalize the dual optimization algorithm of~\cite{abbas2021fastdog} to allow for a larger space of parameter updates. 
\item We make our dual optimization algorithm efficiently differentiable and embed it as a layer in a neural network.
This enables us to predict parameters of the algorithm leading to faster convergence compared to manually designed rules.
\item
We train a predictor for arbitrary non-parametric updates that allow to escape suboptimal fixed points encountered by parametric update steps of~\cite{abbas2021fastdog}.
\item
Our predictors for both of the above updates are trained in a fully unsupervised manner.
Our loss optimizes for producing large improvements in the dual objective.
\item
We show the benefits of our learned approach on a wide range of problems. 
We have chosen structured prediction tasks including graph matching~\cite{kainmueller2014active} and cell tracking~\cite{haller2020primal}.
From theoretical computer science we compare on the QAPLib~\cite{QAPLIB} dataset and on randomly generated independent set problems~\cite{prouvost2020ecole}.
\end{itemize}

\section{Related Work}
\label{sec:related-work}

\subsection{Learning to solve combinatorial optimization}
ML has been used to improve various aspects 
of solving combinatorial problems.
For the standard branch-and-cut ILP solvers the works~\cite{gasse2019exact_co_gnn, gupta2020hybrid_learn_branch, nair2020solving, scavuzzo2022learning_branch_treemdps} learn variable selection for branching.
The approaches~\cite{ding2020accelerating, nair2020solving} learn to fix a subset of integer variables in ILPs to their hopefully optimal values to improve finding high quality primal solutions.
The works~\cite{sonnerat2021learning,wu2021learning} learn variable selection for the large neighborhood search heuristic for obtaining primal solutions to ILPs.
Selecting good cuts through scoring them with neural networks was investigated in~\cite{huang2022learning,turner2022adaptive}.
While all these approaches result in runtime and solution quality improvements, only a few works tackle the important task of speeding up ILP relaxations by ML. Specifically,
the work~\cite{cappart2019improving_dd} used graph neural network (GNN) to predict variable orderings of decision diagrams representing combinatorial optimization problems.
The goal is to obtain an ordering such that a corresponding dual lower bound is maximal.
To our knowledge it is the only work that accelerates ILP relaxation computation with ML.
For constraint satisfaction problems~\cite{selsam2018learning_sat, cameron_2020_e2e_sat,  grohe2021gnn_satisfaction} train GNN while the latter train in an unsupervised manner.
For inference in graphical models~\cite{deng2022deep_bp} learn parameters of belief propagation for faster convergence in a similar spirit to our work. However our method is applicable to a more general class of problems, allows escaping fixed-points, and is scalable to larger problems due to efficient implementation.
For narrow subclasses of problems primal heuristics have been augmented through learning some of their decisions, e.g.\ for capacitated vehicle routing~\cite{nazari2018reinforcement}, graph matching~\cite{wang2021neural_graphmatching} and traveling salesman~\cite{xin2021neurolkh}.
For a more complete overview of ML for combinatorial optimization we refer to the detailed surveys~\cite{bengio2021ml_co_tour, cappart2023combinatorial}.


\subsection{Unrolling algorithms for parameter learning}
Algorithms containing differentiable iterative procedures are combined with neural networks for improving performance of such algorithms. Such approaches show more generalization power than pure neural networks based ones as shown in the survey~\cite{algorithm_unrolling}.
The work of~\cite{lecun_learned_ista} embedded sparse coding algorithms in a neural network by unrolling. For solving inverse problems~\cite{admm_csnet, trainable_diffusion} unroll through ADMM and non-linear diffusion resp. Lastly, neural networks were used to predict update directions for training other neural networks (e.g.\ in~\cite{andrychowicz2016learning_gd}).

\section{Method}
\label{sec:method}

We first recapitulate the Lagrange decomposition approach to binary ILPs from~\cite{lange2021efficient} and generalize the optimization scheme of~\cite{abbas2021fastdog} for faster convergence.
Then we will show how to backpropagate through the optimization scheme allowing to train a graph neural network for predicting its parameters. 

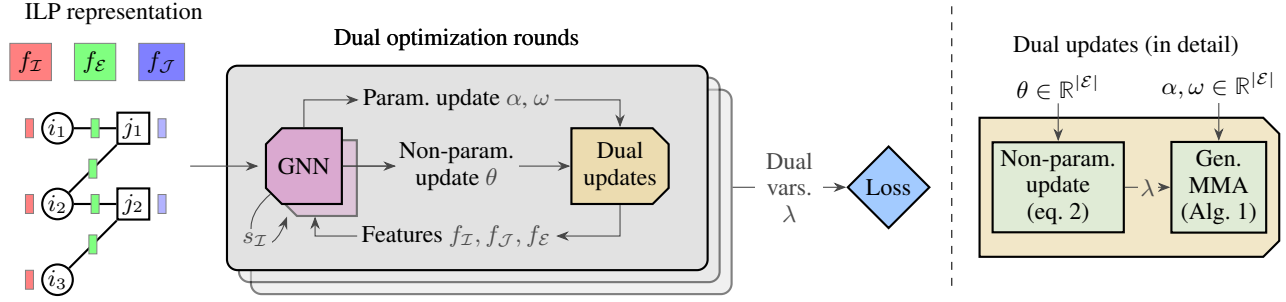
\begin{figure*}[ht]
    \centering
    \begin{tikzpicture}[scale=1.0]
\def\circledarrow#1#2#3{ 
\draw[#1,->, dashed] (#2) +(40:#3) arc(40:-220:#3);
}

\tikzstyle{var_vertex}=[circle, draw, color=black, fill=white, align = center, thick, inner sep = 0.5pt]
\tikzstyle{con_vertex}=[rectangle, draw, color=black, fill=white, align = center, thick, inner sep = 2pt]

\tikzstyle{edge}=[edgecolor!60, line width=0.3mm]
\tikzstyle{feature_node}=[rounded corners=0.2cm, rectangle, draw, color=black, fill=white, densely dotted, inner sep = 2pt]
\tikzstyle{con_f}=[rounded rectangle, draw, color=gray, fill=white, inner sep=1pt, minimum width=1ex]
\tikzstyle{edge_f}=[rounded rectangle, draw, color=gray, fill=white, inner sep=1pt, minimum width=1ex]
\tikzstyle{gnn}=[chamfered rectangle, chamfered rectangle corners={north west,south west}, draw, color=black, fill=gnncolor, thick, minimum height=1cm] 
\tikzstyle{dmma}=[chamfered rectangle, chamfered rectangle corners={north east,south east}, draw, color=black, fill=dmmacolor, thick, minimum height=1cm] 
\tikzstyle{nonparam}=[rectangle, draw, color=black, fill=nonparamcolor!50, align = center, thick, minimum height=1cm]

\tikzset{node distance = 1cm}
\tikzset{>= Stealth}
\tikzstyle{every node}=[font=\small, inner sep = 1.5pt]


\node[style=var_vertex](v1) at (-3.25,0.5) {$i_1$};
\node[style=var_vertex](v2) at (-3.25,-0.5) {$i_2$};
\node[style=var_vertex](v3) at (-3.25,-1.5) {$i_3$};

\node [draw, ultra thin, gray, fill=red!50, rectangle,minimum height = 0.25cm, left=0.1cm of v2] (v2f) {};
\node [draw, ultra thin, gray, fill=red!50, rectangle,minimum height = 0.25cm, left=0.1cm of v1] (v1f) {};
\node [draw, ultra thin, gray, fill=red!50, rectangle,minimum height = 0.25cm, left=0.1cm of v3] (v3f) {};

\node[style=con_vertex](c1) at (-2.25, 0.5) {$j_1$};
\node[style=con_vertex](c2) at (-2.25, -0.5) {$j_2$};

\node [draw, ultra thin, gray, fill=blue!30, rectangle,minimum height = 0.25cm, right=0.1cm of c1] (c1f) {};
\node [draw, ultra thin, gray, fill=blue!30, rectangle,minimum height = 0.25cm, right=0.1cm of c2] (c2f) {};

\draw[style=edge, black, thick] (v1) -- node [draw, ultra thin, gray, rectangle, minimum height = 0.25cm, midway, fill=green!50] (f_e1) {} (c1);
\draw[style=edge, black, thick] (v2) -- node [draw, ultra thin, gray, rectangle, minimum height = 0.25cm, midway, fill=green!50] (f_e) {} (c1);
\draw[style=edge, black, thick] (v2) -- node [draw, ultra thin, gray, rectangle, minimum height = 0.25cm, midway, fill=green!50] (f_e) {} (c2);
\draw[style=edge, black, thick] (v3) -- node [draw, ultra thin, gray, rectangle, minimum height = 0.25cm, midway, fill=green!50] (f_e) {} (c2);

\node[draw=none, above left = 2.1cm and -1.2cm of c2, align = center](graph_text) {\textnormal{ILP  representation}};

\node [draw=gray, ultra thin, fill=green!50, rectangle,minimum height = 0.25cm, above= 0.5cm of f_e1, inner sep = 3pt] (fe) { $f_\mathcal{E}$};
\node [draw=gray, ultra thin, fill=blue!50, rectangle, minimum height = 0.25cm, right = 0.3cm of fe, inner sep = 3pt] (fj) { $f_\mathcal{J}$};
\node [draw=gray, ultra thin, fill=red!50, rectangle,minimum height = 0.25cm, left = 0.3cm of fe, inner sep = 3pt] (fi) { $f_\mathcal{I}$};

\draw[arrowcolor, ->] (-1.5, 0) -- (-0.6, 0);
\node[style=gnn, align = center, copy shadow={opacity=.5, shadow xshift=1.25ex,shadow yshift=-1ex}](gnn1) at (0, 0) {GNN};

\node[draw=none, below right=0.2ex and -2.7ex of gnn1](gnn1_shadow) {};


\node[style=dmma, right=3cm of gnn1, align = center](mma1) {Dual \\ updates};



\draw[arrowcolor, ->] (gnn1.0) -- node [draw=none, fill=bgcolor, midway, align=center] (theta) {\textcolor{black}{\footnotesize{Non-param.}} \\ \textcolor{black}{\footnotesize{update}} $\theta$} (mma1);
\draw[arrowcolor, ->] (gnn1.0) -- ++(0.7,0);





\node[draw=none, above = 0.3cm of theta, align=center,fill=bgcolor](alphaomega){\footnotesize{Param.\ update \textcolor{black!70}{$\alpha$, $\omega$}}};
\draw[->, arrowcolor] (gnn1.90) |- (alphaomega.180);
\draw[->, arrowcolor] (alphaomega.0) -| (mma1.90);


\node[draw=none, below = 0.35cm of theta, align=center](ss_feat){
\footnotesize{Features \textcolor{black!70}{ $f_\mathcal{I}, f_\mathcal{J}, f_\mathcal{E}$}}};
\draw[->, arrowcolor] (mma1.270) |- (ss_feat.0);
\draw[->, arrowcolor] (ss_feat.180) -| (gnn1_shadow.270);


\node[draw=none, black, above = 1.1cm of theta](optround) {\textnormal{Dual optimization rounds}};

\begin{scope}[on background layer]
\node[draw, rounded corners, thick, black, fill=bgcolor, inner xsep=0.5cm, inner ysep=0.25cm, fit=(gnn1) (mma1) (alphaomega) (ss_feat), double copy shadow={opacity=.5, shadow xshift=1ex,shadow yshift=-1ex}] (box_1) {};
\end{scope}


\node[draw=none, below right = 1.75ex of box_1.0](box_1_shadow_right_center){};

\node[diamond, draw=black, inner sep = 2pt, thick, fill=losscolor, align = center, right=1.5cm of box_1_shadow_right_center](loss) {Loss};

\draw[arrowcolor, ->] (box_1_shadow_right_center) -- node [draw=none, fill=white, midway, align = center, inner sep = 4pt] (lambda1) {Dual \\ vars.\ \\ \textcolor{black!70}{$\lambda$}} (loss);

\node[draw=none, black, below = 1.2cm of mma1](repeat) {};

\node[draw=none, below right=0.2ex and -4.9ex of gnn1](gnn1_shadow_lstm) {};
\draw[arrowcolor, in=250, out=220, ->, min distance=0.75cm] (gnn1) to node [draw=none, fill=bgcolor, looseness=10, midway] (alpha1) {$s_\mathcal{I}$} (gnn1_shadow_lstm);

\draw[thin, dashed, black] (current bounding box.south east) ++(8pt,7pt) -- (current bounding box.north east) ++(8pt,10pt);

   
\node[nonparam, right=0.8cm of loss, align=center, inner sep=3pt] (nonp2) {Non-param.\ \\ update \\ (eq.\ 2)};
\node[draw=none, above = 0.5cm of nonp2](theta2) {$\theta \in \mathbb{R}^{\left| \mathcal{E} \right|}$};
\draw[arrowcolor, ->] (theta2) -- (nonp2);

\node[nonparam, right=0.6cm of nonp2, align=center, minimum width = 1.25cm, inner sep=3pt] (mma2) {Gen.\ \\ MMA \\ (Alg.\ 1)};
\node[draw=none, above = 0.5cm  of mma2](mma_params) {$\alpha, \omega \in \mathbb{R}^{\left| \mathcal{E} \right|}$};
\draw[arrowcolor, ->] (mma_params) -- (mma2);

\draw[arrowcolor, ->] (nonp2) -- node [draw=none, fill=dmmacolor!70, midway, align = center, inner sep = 1pt] (lambda2) {\textcolor{black!70}{$\lambda$}} (mma2);


\begin{scope}[on background layer]
\node[dmma, fill=dmmacolor!70, inner ysep=0.2cm, fit=(nonp2) (mma2)] (enclosing) {};
\end{scope}

\node[draw=none, black, above = 1.1cm of theta](optround) {\textnormal{Dual optimization rounds}};

\node[draw=none, black, above left = 1.5cm and -1.4cm of lambda2](dualupdate_expanded) {\textnormal{Dual updates (in detail)}};

\end{tikzpicture}
    \caption[Pipeline]{Our method for optimizing the Lagrangean dual~\eqref{eq:dual-problem}.
    The dual problem is encoded on a bipartite graph containing features $f_\I$, $f_\J$ and $f_\E$ for primal variables, subproblems and dual variables resp.
    A graph neural network (GNN) predicts $\theta, \alpha$, $\omega$ for dual updates.
    In one dual update block (right), current set of Lagrange multipliers $\lambda$ are first updated by the non-parametric update using $\theta$. Afterwards parametric update is done via Alg.~\ref{alg:parallel-mma} using $\alpha$, $\omega$.
    The updated solver features $f$ and LSTM cell states $s_\I$ are sent to the GNN in next optimization round.
    See Sec.~\ref{sec:pipeline} for further details.}
    \label{fig:pipeline}
\end{figure*}

\subsection{Lagrange Decomposition}

\begin{defn}[Binary Program~\cite{lange2021efficient}]
Let a linear objective $c \in \R^n$ and  $m$ variable subsets $\I_j \subset [n]$ of constraints with feasible set $\X_j \subset \{0,1\}^{\I_j}$ for $j \in [m]$ be given.
The ensuing binary program is 
\begin{equation}
\label{eq:binary-program}
    \min_{x \in \{0,1\}^n} \la c, x \ra \quad \text{s.t.} \quad x_{\I_j} \in \X_j \quad \forall j \in [m]\,,
\tag{BP} 
\end{equation}
where $x_{\I_j}$ is the restriction to variables in $\I_j$.
\end{defn}

Any binary ILP $\min_{x \in \{0,1\}^n} \la c,x\ra \text{ s.t. } Ax \leq b$ where $A \in \R^{m \times n}$ can be written as~\eqref{eq:binary-program} by associating each constraint $a_{j}^T x \leq b_j$ for $j \in [m]$ with its own subproblem $\X_j$. 
In order to obtain a problem formulation amenable for parallel optimization we consider its Lagrange dual which decomposes the full problem~\eqref{eq:binary-program} into a series of coupled subproblems.

\begin{defn}[Lagrangean dual problem~\cite{lange2021efficient}]

Define the set of subproblems that constrain variable $i$ as $\J_i = \{j \in [m] \mid i \in \I_j\}$. Let the energy for subproblem $j \in [m]$ w.r.t.\ Lagrange variables $\lambda_{\bullet j} = (\lambda_{ij})_{i\in\I_j} \in \R^{\I_j}$ be
\begin{equation}
E^j(\lambda_{\bullet j}) = \min_{x \in \X_j} \la  \lambda_{\bullet j}, x \ra \,.
\label{eq:dual_subproblem}
\end{equation}
Then the Lagrangean dual problem is defined as
\begin{align}
\max_\lambda \sum_{j \in [m]} E^j(\lambda_{\bullet j}) \quad
\text{s.t.} \quad \sum_{j \in \J_i} \lambda_{ij} = c_i \quad \forall i \in [n]. \label{eq:dual-problem} \tag{D}
\end{align}

\end{defn}
The problem~\eqref{eq:dual-problem} provides a lower bound to the NP-Hard optimization problem~\eqref{eq:binary-program} and is also useful for primal recovery~\cite{abbas2021fastdog}. Our goal is to learn a neural network for optimizing the dual~\eqref{eq:dual-problem} efficiently and to reach better objective values.

\subsection{Optimization of Lagrangean dual}
The work of~\citet{abbas2021fastdog} proposed a parallelization friendly iterative scheme for optimizing~\eqref{eq:dual-problem} with hand-designed parameters. We generalize their scheme in Algorithm~\ref{alg:parallel-mma}, exposing a much larger set of parameters allowing more control over the optimization process. Since this large parameter space is difficult to be tuned manually, we will employ a GNN for predicting these parameters.

In detail, Alg.~\ref{alg:parallel-mma} greedily assigns the Lagrange variables in $u$-many disjoint blocks $B_1, \dotsc, B_u$ in such a way that each block contains at most one Lagrange variable from each subproblem and all variables within a block are updated in parallel (same as~\cite{abbas2021fastdog}). 
The dual update scheme relies on computing min-marginal differences i.e., the difference of subproblem objectives when a certain variable is set to $1$ minus its objective when the same variable is set to $0$ (line~\ref{alg:min-marginal-difference}).
These min-marginal differences are averaged out across subproblems via updates to Lagrange variables (line~\ref{alg:lambda-update}). Our algorithm relies on two important set of parameters, damping factors and averaging weights. The damping factor $\omega_{ij}$ determine the fraction of min-marginal difference to subtract from variable $i$ in subproblem $j$. The averaging weights $\alpha_{ij}$ parameterize the fraction of total min-marginal difference $\sum_{ik} M_{ik}$ variable $i$ in subproblem $j$ receives. 

\begin{rem}
    The deferred min-marginal averaging algorithm of~\cite{abbas2021fastdog} is a specialized form of our generalized Algorithm~\ref{alg:parallel-mma} if the parameters were set as $\omega_{ij} = 0.5$ and $\alpha_{ij} = \nicefrac{1}{\abs{\J_i}}$ for all $i, j$.
\end{rem}

We generalize the min-marginal update step by considering damping factors to be in $(0,1)$ and averaging weights to be arbitrary convex combinations. We show that this generalized update step still preserves the desirable property of guaranteed non-improvement in the dual objective.



\begin{prop}[Dual Feasibility and Monotonicity of Generalized Min-marginal Averaging]
For any $\alpha_{ij} \geq 0$ with $\sum_{j \in \J_i} \alpha_{ij} = 1$ and $\omega_{ij} \in [0,1]$ the min-marginal averaging step in line~\ref{alg:lambda-update} in Algorithm~\ref{alg:parallel-mma} retains dual feasibility and is non-decreasing in the dual lower bound.
\label{prop:dual-feasible-monotone}
\end{prop}
\begin{algorithm}[h]
\SetKw{Break}{break}
\SetKwFunction{proc}{BlockUpdate}
\KwInput{
Lagrange variables $\lambda_{ij} \,  \forall i \in [n], j \in \J_i$,\\
\hspace{22pt} damping factors $\omega_{ij} \in (0,1) \, \forall i \in [n], j \in \J_i$,\\
\hspace{22pt} averaging weights $\alpha_{ij} \in (0,1) \, \forall i \in [n], j \in \J_i$,\\
\hspace{22pt} max. number of iterations $T$.
} 
Initialize deferred min-marginal diff.  $M = \0$\;
\label{alg:initialization-line}
\For{$T$ iterations}
{
\For{block $B \in (B_1, \dotsc B_u)$}{
    $\lambda, M \leftarrow$ \proc($B, \lambda, M, \alpha, \omega$)\;
    \label{alg:after-forward-block-update}
}
\For{block $B \in (B_u, \dotsc B_1)$}{
    $\lambda, M \leftarrow$ \proc($B, \lambda, M, \alpha, \omega$)\;
    \label{alg:after-backward-block-update}
}
\label{alg:optimization-end}
}
\Return{$\lambda$, $M$}
\caption{Generalized Min-Marginal Averaging}
\label{alg:parallel-mma}

\SetKwProg{myproc}{Procedure}{}{}
\myproc{\proc{$B, \lambda^{\old}, M^{\old}, \alpha, \omega$}}{
    \label{alg:parallel-mma-half-iter-start}
    \label{alg:variable-order-loop-start}
    \For{$ij \in B$ in parallel}
    {
    \label{alg:min-marginal-computation-start}
    $M_{ij}^{\new} = \omega_{ij} \big[ \min\limits_{\substack{x \in \X_j :\\ x_i = 1}} \la  \lambda_{\bullet j}^\old, x \ra - \min\limits_{\substack{x \in \X_j :\\ x_i = 0}} \la \lambda_{\bullet j}^\old, x \ra \big]$\;
    \label{alg:min-marginal-difference}
    \label{alg:min-marginal-computation-end}%
    \smallskip
    $\lambda_{ij}^{\new} = \lambda_{ij}^{\old} - M_{ij}^{\new} + \alpha_{ij} \sum_{k \in \J_i} M_{ik}^{\old}$\;
    \label{alg:lambda-update}
    }
    \label{alg:deferred-min-marginals-update}
    \Return{$\lambda^{\new}$, $M^{\new}$}
    \label{alg:parallel-mma-half-iter-end}
}
\end{algorithm}

\subsection{Backpropagation through dual optimization}

We show below how to differentiate through Algorithm~\ref{alg:parallel-mma} with respect to its parameters $\alpha$ and $\omega$. This will ultimately allow us to learn these parameters such that faster convergence is achieved.
To this end we describe backpropagation for a block update (lines~\ref{alg:parallel-mma-half-iter-start}-\ref{alg:parallel-mma-half-iter-end}) of Alg.~\ref{alg:parallel-mma}.
All other operations can be tackled by automatic differentiation.
For a block $B$ in $\{B_1, \dotsc, B_u\}$ we view the Lagrangean update as a mapping 
$\mathcal{H} : (\R^{\lvert B \rvert })^4 \rightarrow (\R^{\lvert B \rvert})^2$, 
$(\lambda^{\old},M^{\old},\alpha,\omega) \mapsto (\lambda^{\new},M^{\new})$.

Given a loss function $\mathcal{L}: \R^{N} \rightarrow \R$ we denote ${\partial \mathcal{L}}/ {\partial x}$ by $\der{x}$. Algorithm~\ref{alg:parallel-mma-backprop} shows backpropagation through $\mathcal{H}$ to compute the gradients
$\der{\lambda}^\old$,
$\der{M}^\old$,
$\der{\alpha}$ and
$\der{\omega}$.

\begin{algorithm}
\SetInd{3pt}{6pt}
\KwInput{
Forward pass inputs: $B, \lambda^{\old}, M^{\old}, \alpha, \omega$, gradients of forward pass output: $\der{\lambda}^{\new}$, $\der{M}^{\new}$, gradients of parameters $\der{\alpha}, \der{\omega}$
} 
\For{$ij \in B$ in parallel}
{
\label{alg:backprop-variable-order-loop-start}
$\der{M}_{ij}^\old = \sum_{k \in \J_i} \der{\lambda}_{ik}^\new \alpha_{ik}$ \;
$\der{M}_{ij}^\new = \der{M}_{ij}^\new -\der{\lambda}_{ij}^\new$ \;
\label{alg:backprop-acc-min-marginals-end}
$\der{\alpha}_{ij} = \der{\alpha}_{ij} + \der{\lambda}_{ij} \sum_{k \in \J_i}M_{ik}^\old$ \;
$\der{\omega}_{ij} = \der{\omega}_{ij} + \der{M}_{ij}^\new [M_{ij}^\new/\omega_{ij}]$ \;
\label{alg:backprop-acc-params-end}
Compute minimizers for $\beta \in \{0, 1\}$\, $s^j(i, \beta) = \argmin\limits_{x \in \X_j: x_i = \beta} \la  \lambda_{\bullet j}^\old, x \ra$ \;
\label{alg:backprop-compute-minimizers}
$\der{\lambda}_{pj}^\old = \der{\lambda}_{pj}^\new +
\der{M}_{ij}^\new \omega_{ij} [s_p^j(i, 1) - s_p^j(i, 0)], \, \forall p \in \I_j$\;   
\label{alg:backprop-acc-minimizers}
}
\Return{$\der{\lambda}^\old, \der{M}^\old, \der{\alpha}, \der{\omega}$}
\caption{\texttt{BlockUpdate} backpropagation}
\label{alg:parallel-mma-backprop}
\end{algorithm}

\begin{samepage}
\begin{prop}
Alg.~\ref{alg:parallel-mma-backprop} performs backprop.\ through $\mathcal{H}$.
\label{prop:half-iteration-grads}
\end{prop}
\end{samepage}
\paragraph{Efficient Implementation}
Generally, the naive computation of min-marginal differences and its backpropagation are both expensive operations as they require solving two optimization problems for each dual variable.
The works of~\citet{abbas2021fastdog, lange2021efficient} represent each subproblem by a binary decision diagram (BDD) for fast computation of min-marginal differences.
Their algorithm results in a computation graph involving only elementary arithmetic operations and taking minima over several variables.
Using this computational graph we can implement the abstract Algorithm~\ref{alg:parallel-mma-backprop} efficiently on GPU. For further performance gains we implement custom backpropagation routines in CUDA for more than an order of magnitude decrease in runtime and memory usage as shown in Table 4 of the Appendix.
\subsection{Non-Parametric Update Steps}
Although the min-marginal averaging scheme of Alg.~\ref{alg:parallel-mma} guarantees a non-decreasing lower bound, it can get stuck in suboptimal fixed points, see~\cite{werner2007linear} for a discussion for the special case of MAP inference in Markov Random Fields and~\cite{werner2019relative} for a more general setting.
To address this issue we allow arbitrary updates to Lagrange variables through a vector $\hat{\theta} \in \R^{\lvert \lambda \vert}$ as
\begin{equation}
    \lambda_{ij} \leftarrow \lambda_{ij} + \hat{\theta}_{ij} - \frac{1}{\lvert \J_i \rvert}\sum_{k \in \J_i} \hat{\theta}_{ik}, \, \forall i \in [n], j \in \J_i,
    \label{eq:non-parametric-update}
\end{equation}
where the last term ensures feasibility of updated Lagrange variables w.r.t.\ the dual problem~\eqref{eq:dual-problem}. 

\subsection{Graph Neural Network}
We train a graph neural network (GNN) to predict the parameters $\alpha, \omega \in \R^{\lvert \lambda \vert}$ of Alg.~\ref{alg:parallel-mma} and also the non-parametric update $\theta \in \R^{\lvert \lambda \vert}$ for~\eqref{eq:non-parametric-update}. 
To this end we encode the dual problem~\eqref{eq:dual-problem} on a bipartite graph $\mathcal{G} = (\mathcal{V}, \mathcal{E})$.
Its nodes correspond to primal variables $\I$ and subproblems $\J$ i.e., $\mathcal{V} = \I \cup \J$ and edges $\E = \{ij \mid i \in \I, j \in \J_i\}$ correspond to Lagrange multipliers.
We need to predict values of $\alpha_{ij}$, $\omega_{ij}$ and $\theta_{ij}$ for each edge $ij$ in $\E$. 
We associate features $f = (f_\I, f_\J, f_\E)$ with each entity (nodes, edges) of the bipartite graph. Lagrange multipliers $\lambda^{\old}$ and deferred min-marginals $M^{\old}$ encode the current state of Alg.~\ref{alg:parallel-mma} as a part of edge features. Additionally, we encode a number of quantities as features which can allow the GNN to make better updates.
Specifically, a subgradient of the dual problem~\eqref{eq:dual-problem} is encoded in the edge features $f_\E$ and a history of previous dual objectives for each subproblem is encoded in the constraint features $f_\J$. This enables our GNN to effectively utilize more information in parameter prediction than conventional hand-designed updates rules can manage. For example~\cite{abbas2021fastdog} can get stuck in suboptimal fixed points due to zero min-marginal differences~\cite{werner2019relative}. Since the GNN additionally has access to subgradient of the dual problem it can escape such fixed points. 
A complete list of features is provided in the Appendix. 

\paragraph{Graph convolution}
We use the transformer based graph convolution scheme~\cite{transformerconv}. 
We first compute embeddings of all subproblems $j$ in $\J$ by receiving messages from adjacent nodes and edges as
\begin{equation}
\begin{aligned}
     \MP_\J&(f_\I, f_\J, f_\E, \E)_j = \mathbf{W_s} f_j \, + \\
    & \sum_{i \mid ij \in \E} a_{ij}(f_j, f_\I, f_\E; \mathbf{W_a}) \left[ \mathbf{W_t} f_{i} + \mathbf{W_e}f_{ij}\right],
    \label{eq:graph-conv}
\end{aligned}
\end{equation}
where $\mathbf{W_a}, \mathbf{W_s}, \mathbf{W_t}, \mathbf{W_e}$ are trainable parameters and $a_{ij}(f_j, f_\I, f_\E; \mathbf{W_a})$ is the softmax attention weight between nodes $i$ and $j$ parameterized by $\mathbf{W_a}$.
Afterwards we perform message passing in the reverse direction to compute embeddings for variables $\I$. A similar strategy for message passing on bipartite graphs was followed in~\cite{gasse2019exact_co_gnn}.

\myparagraph{Recurrent connections}
Our default GNN as mentioned above only uses hand-crafted features to maintain a history of previous optimization rounds. To learn a summary of the past updates we optionally allow recurrent connections through an LSTM with forget gate~\cite{lstm1999}. The LSTM is only applied on primal variable nodes $\I$ and maintains cell states $s_\I$ which can be updated and used for parameter prediction in subsequent optimization rounds.

\myparagraph{Prediction}
The learned embeddings from GNN, LSTM outputs and solver features are consumed by a multi-layer perceptron $\Phi$ to predict the required variables for each edge $ij$ in $\E$. Afterwards we transform these outputs so that they satisfy Prop.~\ref{prop:dual-feasible-monotone}.
The exact sequence of operations performed by the graph neural network are shown in Alg.~\ref{alg:gnn_message_passing} where $[u_1, \dotsc, u_k]$ denotes concatenation of vectors $u_1, \dotsc, u_k$, $\texttt{LN}$ denotes layer normalization~\cite{ba2016layernorm} and \texttt{LSTM$_\I$} stands for an LSTM cell operating on primal variables $\I$.

\begin{algorithm}
\newcommand\mycommfont[1]{\footnotesize\ttfamily\textcolor{blue}{#1}}
\SetCommentSty{mycommfont}
\KwInput{
Primal variable features $f_\I$ and cell states $s_\I$, 
Subproblem features $f_\J$, 
Dual variable (edge) features $f_\E$,
Set of edges $\E$.
}

\tcp{Compute subproblems embeddings}
$h_\J = \texttt{ReLU}\left(\LN\left(\MP_\J\left(f_\I, f_\J, f_\E, \E \right)\right)\right)$\;
\tcp{Compute primal var embeddings}
$h_\I = \texttt{ReLU}\left(\LN\left(\MP_\I\left(f_\I, [f_\J, h_\J], f_\E, \E \right)\right)\right)$
\label{alg:mp_primal_update}

\tcp{Compute output and cell state}
$z_\I, s_\I = \texttt{LSTM}_\I(h_\I, s_\I)$

\tcp{Prediction per edge}
$(\hat{\alpha}, \hat{\omega}, \hat{\theta}) = \Phi\left([f_\I, h_\I, z_\I], [f_\J, h_\J], f_\E, \E \right)$


\tcp{Ensure non-decreasing obj., Prop~\ref{prop:dual-feasible-monotone}:}
$\alpha_{i \bullet} = \texttt{Softmax}(\hat{\alpha}_{i \bullet})$, $\forall i \in \I$,\,
$\omega = \texttt{Sigmoid}(\hat{\omega})$

\tcp{Maintain dual feasibility}
$\theta_{i \bullet} = \hat{\theta}_{i \bullet} - \frac{1}{\lvert \J_i \rvert}\sum_{k \in \J_i} \hat{\theta}_{ik}$, $\forall i \in \I$ 

\Return{$\alpha, \omega, \theta, s_\I$}
\caption{Parameter prediction by GNN}
\label{alg:gnn_message_passing}
\end{algorithm}

\paragraph{Loss}
Given the Lagrange variables $\lambda$ we directly use the dual objective~\eqref{eq:dual-problem} as an unsupervised loss to train the GNN. Thus, we maximize the loss $L$ defined as
\begin{equation}
    \mathcal{L}(\lambda) = \sum_{j \in [m]} E^j(\lambda_{\bullet j}).
    \label{eq:loss}
\end{equation}
For a mini-batch of instances during training we take the mean of corresponding per-instance losses. For backpropagation, gradient of the loss $\mathcal{L}$ w.r.t.\ Lagrange variables of a subproblem $j$ is computed by finding a minimizing assignment for that subproblem, written as
$
\left(\frac{\partial \mathcal{L}}{\partial \lambda}\right)_{\bullet j} = \argmin_{x \in \X_j} \la  \lambda_{\bullet j}, x \ra \in \{0, 1\}^{\I_j}
$.
This gradient is then sent as input for backpropagation. For computing the minimizing assignment efficiently we use binary decision diagram representation of each subproblem as in~\cite{abbas2021fastdog, lange2021efficient}.

\subsection{Overall pipeline}
\label{sec:pipeline}

We train our pipeline (Fig.~\ref{fig:pipeline}) which contains multiple dual optimization rounds in a fashion similar to that of recurrent neural networks. One round of our dual optimization consists of message passing by GNN, a non-parametric update step and $T$ iterations of generalized min-marginal averaging. For computational efficiency we run our pipeline for at most $R$ dual optimization rounds during training.
On each mini-batch we randomly sample a number of optimization rounds $r$ in $[R]$, run $r - 1$ rounds without tracking gradients and backpropagate through the last round by computing the loss~\eqref{eq:loss}. For the pipeline with recurrent connections we backpropagate through last $3$ rounds and apply the loss after each of these rounds. 
Since the task of dual optimization is relatively easier in early rounds as compared to later ones we use two neural networks.
The early stage network is trained if the randomly sampled $r$ is in $[0, R/2]$ and the late stage network is chosen otherwise.
During testing we switch to the later stage network when the relative improvement in the dual objective by the early stage network becomes less than $10^{-6}$. For computational efficiency during testing we query the GNN for parameter updates only after $T \gg 1$ iterations of Alg.~\ref{alg:parallel-mma}.

\section{Experiments}
\label{sec:experiments}

\begin{table*}
    \caption{Results on tests instances where the values are averaged within a dataset. Numbers in bold highlight the best performance and underlines indicate the second best objective.}
    \label{tab:combined-results}
    \centering
    \begin{tabular}{l rrr rrr rrr rrr}
    \toprule
    \multirow{2}{*}{} & \multicolumn{3}{c}{\textit{Cell tracking}} & \multicolumn{3}{c}{\textit{Graph matching}} &  \multicolumn{3}{c}{\textit{Independent set}} & \multicolumn{3}{c}{\textit{QAPLib}} \\
    \cmidrule(lr){2-4} \cmidrule(lr){5-7} \cmidrule(lr){8-10} \cmidrule(lr){11-13}
     & $g_I$ & $E(\times 10^8)$ & $t[s]$
     & $g_I$ & $E(\times 10^4)$ & $t[s]$
     & $g_I$ & $E(\times 10^4)$ & $t[s]$
     & $g_I$ & $E(\times 10^6)$ & $t[s]$
     \\
    \midrule
    \texttt{Gurobi} 
    & $18$ & $-\textbf{3.852}$ & $809$ 
    & $9$ & $-\textbf{4.8433}$ & $278$ 
    & $14$ & $-\textbf{2.4457}$ & $52$ 
    & $3472$ & ${0.9}$ & $2618$ 
    \\
    \texttt{Spec.}
    & - & $-3.866$ & $1673$ &
    - & $-4.8443$ & $100$ &
    - & - & - &
    - & - & -
    \\
    \texttt{FastDOG}
    & $7$ & $-3.863$ & $1005$ 
    & $21$ & $-4.8912$ & $61$ 
    & $42$ & $-2.4913$ & $9$ 
    & $276$ & $5.7$ & $1680$ 
    \\
    \texttt{DOGE}
    & ${2.4}$ & $-3.854$ & $1015$  
    & ${0.3}$ & $-4.8439$ & $17$ 
    & ${0.3}$ & $-2.4460$ & $8$ 
    & $320$ & \underline{$12.1$} & $720$ 
    \\
    \texttt{DOGE-M}
    & $\textbf{2.1}$ & \underline{$-3.854$} & $730$  
    & $\textbf{0.2}$ & \underline{$-4.8436$} & $21$ 
    & $\textbf{0.2}$ & \underline{$-2.4459$} & $5$ 
    & $\textbf{131}$ & $\textbf{14.5}$ & $861$ 
    \\
    \bottomrule
    \end{tabular}
\end{table*}

\begin{table*}
\caption{Ablation study results on the \textit{Graph matching} dataset.
w/o GNN: Use only the two predictors $\Phi$ without GNN for early and late stage optimization;
same network: use one network (GNN, $\Phi$) for both early and late stage;
only non-param., param.: predict only the non-parametric update~\eqref{eq:non-parametric-update} or the parametric update (Alg.~\ref{alg:parallel-mma});
w/o $\alpha$, $\omega$: does not predict $\alpha$ or $\omega$ resp.}
\label{tab:ablation}
\centering
    \begin{tabular}{c c c c c c c c c c} 
        \toprule  
         & \multirow{2}{*}{\parbox{1.3cm}{\centering w/o learn. \\ (\texttt{FastDOG})}} &
        \multirow{2}{*}{\parbox{1.0cm}{\centering w/o \\ GNN}} &
        \multirow{2}{*}{\parbox{1.0cm}{\centering same \\ network}} &
        \multirow{2}{*}{\parbox{1.7cm}{\centering only \\ non-param.}} & \multirow{2}{*}{\parbox{1cm}{\centering only \\ param.}} & 
        \multirow{2}{*}{w/o $\alpha$} & \multirow{2}{*}{w/o $\omega$} &
        \multirow{2}{*}{\texttt{DOGE}} & 
        \multirow{2}{*}{\texttt{DOGE-M}} \\ 
        & & & & & & & & & \\
        \midrule
        $g_I$ $(\downarrow)$ & $21$ 
        & $0.42$ & $0.95$ 
        & $2.3$ & $0.7$ 
        & $0.36$ & ${0.35}$
        & ${0.33}$ & $\textbf{0.19}$ \\
        \midrule
        $E$ $(\uparrow) $ & $-48912$ 
        & $-48440$ & $-48444$
        & $-48476$ & $-48444$
        & $-{48439}$ & $-{48439}$ 
        & $-{48439}$ & $-\textbf{48436}$ \\
        \midrule
        $t[s]$ $(\downarrow)$ & $61$ 
        & $29$ &  $24$
        & $51$ & $74$
        & $30$ & $30$ 
        & $17$ & $21$ \\
        \bottomrule
    \end{tabular}
\end{table*}

\begin{figure*}
\centering
\includegraphics[width=0.8\textwidth]{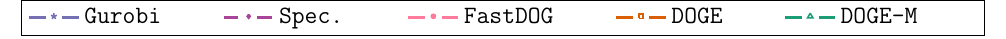}
\subcaptionbox{\textit{Cell tracking}} {\includegraphics[width=0.51\textwidth]{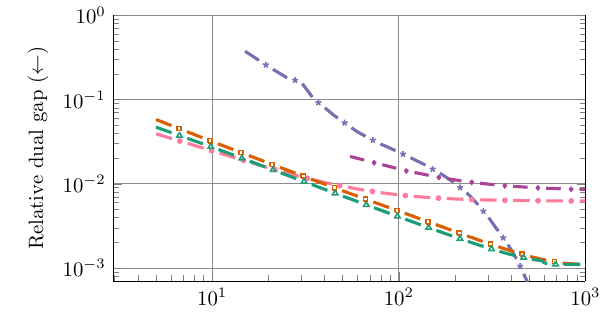}}%
\label{fig:convergence_plot_ct}\hfill%
\subcaptionbox{\textit{Graph matching}} {\includegraphics[width=0.485\textwidth]{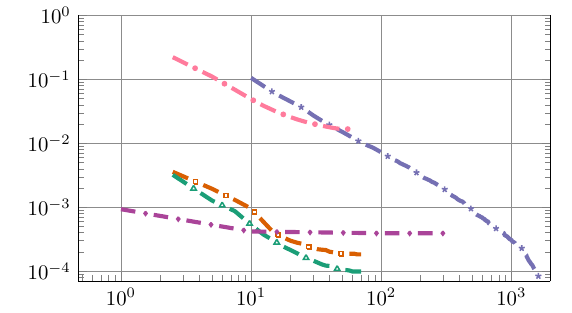}}
\label{fig:convergence_plot_worms}
\subcaptionbox{\textit{Independent set}} {\includegraphics[width=0.51\textwidth]{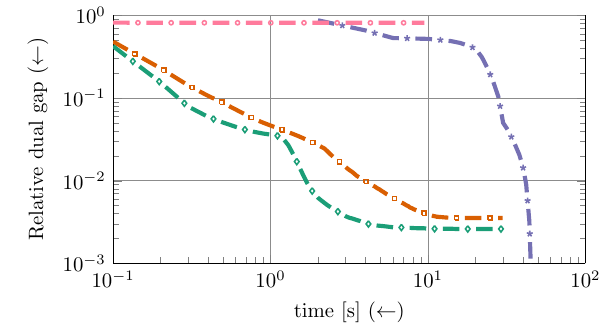}}%
\label{fig:convergence_plot_ind_set}\hfill%
\subcaptionbox{\textit{QAPLib}} {\includegraphics[width=0.49\textwidth]{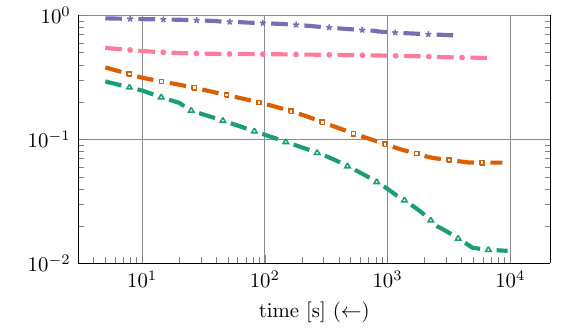}}%
\label{fig:convergence_plot_qaplib}

  \caption{
  Convergence plots for $g(t)$ the relative dual gap to the optimum (or maximum suboptimal objective among all methods) of the relaxation~\eqref{eq:dual-problem}. X-axis indicates wall clock time and both axes are logarithmic. The value of $g(t)$ is averaged over all test instances in each dataset.
  }
  \label{fig:convergence_plots} 
\end{figure*}

\subsection{Evaluation}
As main evaluation metric we report convergence plots of the relative dual gap $g(t) \in [0,1]$ at time $t$ by 
    $
    g(t) = \min\left(\frac{d^* - d(t)}{d^* - d_{init}}, 1.0\right)
    $
where $d(t)$ is the dual objective at time $t$, $d^*$ is the optimal (or best known) objective value of the Lagrange relaxation~\eqref{eq:dual-problem} and $d_{init}$ is the objective before optimization as computed by~\cite{abbas2021fastdog}.
Additionally we also report per dataset averages of best objective value ($E$), time taken ($t$) to obtain best objective and relative dual gap integral $g_I = \int g(t)dt$~\cite{berthold_primal_heuristics}. The latter metric $g_I$ allows to measure quality of the solution in conjunction with time take to obtain this solution. 

\subsection{Datasets}
\label{subsec:datasets}
We evaluate our approach on a variety of datasets from different domains. For each dataset we train our pipeline on smaller instances and test on larger ones.  
\begin{description}[itemsep=2pt,parsep=0pt]
\item[\textnormal{\textit{Cell tracking (CT)}}:] Instances of developing flywing tissue from cell tracking challenge~\cite{ulman2017objective_comp_ct} processed by~\cite{haller2020primal} and obtained from~\cite{swoboda2022structured}. We use the largest and hardest 3 instances, train on the 2 smaller instances and test on the largest one.

\item[\textnormal{\textit{Graph matching (GM)}}:] Instances of graph matching for matching nuclei in 3D microscopic images~\cite{long20093d_worms} processed by~\cite{kainmueller2014active} and made publicly available through~\cite{swoboda2022structured} as ILPs.  We train on $10$ instances and test on the remaining $20$. 

\item[\textnormal{\textit{Independent set (IS)}}:] Random instances of independent set problem generated using~\cite{prouvost2020ecole}.
For training we use $240$ instances with $10k$ nodes each and test on $60$ instances with $50k$ nodes. 

\item[\textnormal{\textit{QAPLib}}:] The benchmark dataset for quadratic assignment problems used in the combinatorial optimization community~\cite{QAPLIB}. The benchmark contains problems arising from a variety of domains e.g., keyboard design, hospital layout, circuit synthesis, facility location etc. We train on $61$ instances having up to $0.6M$ Lagrange variables and test on $35$ instances having up to $10.6M$ Lagrange variables. Conversion to ILP is done via (2.7)-(2.16) of~\cite{loiola2007survey}
\end{description}

For each dataset the size of problems~\eqref{eq:dual-problem} are reported in Table~\ref{tab:dataset_stats}. Due to varying instance sizes we use a separate set of hyperparameters for each dataset given in Table 5 of the Appendix\footnote{Appendix in our full paper~\cite{abbas2022dogetrain_arxiv}.}. 
For the \textit{CT} dataset we only predict $\theta \in \R^{\abs{\lambda}}$ for non-parametric update steps~\eqref{eq:non-parametric-update} and fix the parameters $\alpha, \omega$ in Alg.~\ref{alg:parallel-mma} to their default values from~\cite{abbas2021fastdog}. Learning these parameters gave slightly worse training loss at convergence possibly due to small training set size.

\begin{table}[H]
    \caption{
    Statistics of datasets where the values are averaged within each train/test split.  Number of edges in the GNN equal the number of Lagrange multipliers $\lambda$. 
    }
    \label{tab:dataset_stats}
    \centering
    \resizebox{1\linewidth}{!}{
    \begin{tabular}{l ccccccc}
    \toprule
        \multirow{2}{*}{Dataset} & 
        \multicolumn{2}{c}{\# variables ($\times 10^6$)} &
        \multicolumn{2}{c}{\# constraints ($\times 10^6$)} &
        \multicolumn{2}{c}{\# edges  ($\times 10^6$)} \\
        \cmidrule(lr){2-3} \cmidrule(lr){4-5} \cmidrule(lr){6-7}
        & train & test & train & test & train & test \\
    \midrule
        \textit{CT} & $3.1$ & $10.1$& $0.6$ & $2.2$ & $8.5$ & $27.5$ \\
        \textit{GM} & $1.5$ & $0.1$ & $1.5$ & $0.1$ & $3.3$ & $3.3$ \\
        \textit{IS} & $0.01$ & $0.05$ & $0.04$ & $0.4$ & $0.1$ & $1.1$ \\
        \textit{QAPLib} & $0.1$ & $2.5$ & $0.02$ & $0.2$ & $0.6$ & $10.6$ \\
    \bottomrule
\end{tabular}
    }
\end{table}

\subsection{Algorithms}
\begin{description}[itemsep=2pt,parsep=0pt] 
	\item[\normalfont\texttt{Gurobi}:] The dual simplex algorithm from the commercial solver~\cite{gurobi}.
	\item[\normalfont\texttt{Spec.}:] For graph matching and cell tracking datasets we also report results of state-of-the-art dataset specific solvers. For cell tracking the solver of~\cite{haller2020primal} and for graph matching the best performing solver (fm-bca) from recent benchmark~\cite{haller2022gm_benchmark}.
	\item[\normalfont\texttt{FastDOG}:] The non-learned baseline~\cite{abbas2021fastdog} with their hand designed parameters $\omega_{ij} = 0.5$ and $\alpha_{ij} = {1} / {\abs{\J_i}}$ as a specialization of Alg.~\ref{alg:parallel-mma}.
	\item[\normalfont\texttt{DOGE}:] Our approach where we learn to predict parametric and non-parametric updates by using two graph neural networks for early and late-stage optimization. Size of the learned embeddings $h$ computed by the GNN in Alg.~\ref{alg:gnn_message_passing} is set to $16$ for nodes and $8$ for edges.
	For computing attention weights in~\eqref{eq:graph-conv} we use only one attention head for efficiency.
	The predictor $\Phi$ in Alg.~\ref{alg:gnn_message_passing} contains $4$ linear layers with the ReLU activation. We train the networks using the Adam optimizer~\cite{kingma2014adam}. To prevent gradient overflow we use gradient clipping on model parameters by an $l^2$ norm of $50$. The number of trainable parameters is $8k$.
	\item[\normalfont\texttt{DOGE-M}:] Variant of our method where we additionally use recurrent connections using LSTM. The cell state vector $s_i$ for each primal variable node $i \in \I$ has a size of 16. The number of trainable parameters is $12k$.
\end{description}
Note the test instances require millions of solver parameters to be predicted (ref.\ Table~\ref{tab:dataset_stats}) while our largest GNN has $12k$ parameters.

For training we use PyTorch and implement the Algorithms~\ref{alg:parallel-mma} and~\ref{alg:parallel-mma-backprop} in CUDA. CPU solvers use AMD EPYC 7702 CPU with 16 threads. GPU solvers use either an NVIDIA RTX 8000 (48GB) or a A100 (80GB) GPU depending on problem size.

\subsection{Results}
\label{subsec:results}
For each dataset we evaluate our methods on corresponding testing split.
Convergence plots of relative dual gaps change (averaged over all test instances) are given in Figure~\ref{fig:convergence_plots}.
Other evaluation metrics are reported in Table~\ref{tab:combined-results}. For further details we refer to the Appendix. 

\paragraph{Discussion}
As compared to the non-learned baseline \normalfont{\texttt{FastDOG}} we reach an order of magnitude more accurate relaxation solutions, almost closing the gap to optimum as computed by \normalfont{\texttt{Gurobi}}. Even though given unlimited time \normalfont{\texttt{Gurobi}} attains the optimum, we reach reasonably close values that are considered correct for practical purposes. 
For example the graph matching benchmark~\cite{haller2022gm_benchmark} considers a relative gap less than $10^{-3}$ as optimal (we achieve $10^{-4}$).
Moreover our learned solvers reach much better objective values as compared to specialized solvers.
Using LSTM in \normalfont{\texttt{DOGE-M}} further improves the performance especially on the most difficult \textit{QAPLib} dataset.
On \textit{QAPLib} \normalfont{\texttt{Gurobi}} does not converge on instances with more than 40 nodes within the time limit of one hour.
We show convergence plots for smaller instances in the Appendix.
The difference to \normalfont{\texttt{Gurobi}} is most pronounced w.r.t.\ anytime performance measured by $g_I$, as our solver reaches good solutions relatively early.


\paragraph{Ablation study}
We evaluate the importance of various components in our approach. Starting from~\cite{abbas2021fastdog} as a baseline we first predict all parameters $\alpha, \omega, \theta$ through the two multi-layer perceptrons $\Phi$ for early and late stage optimization without using GNN. Next, we report results of using one network (instead of two) which is trained and tested for both early and later rounds of dual optimization. Lastly, we aim to seek the importance of learning parameters of Alg.~\ref{alg:parallel-mma-backprop} and the non-parametric update~\eqref{eq:non-parametric-update}. To this end, we learn to predict only the non-parametric update and apply the loss directly on updated $\lambda$ without requiring backpropagation through Alg.~\ref{alg:parallel-mma}. We also try learning a subset of parameters i.e., not predicting averaging weights $\alpha$ or damping factors $\omega$. Lastly, we report results of \normalfont{\texttt{DOGE-M}} which uses recurrent connections. The results for \textit{graph matching} dataset are in Table~\ref{tab:ablation}. Results on other datasets are provided in the Appendix.  

Firstly, from our ablation study we observe that learning even one of the two types of updates i.e., non-parametric or parametric already gives better results than the non-learned solver \texttt{FastDOG}.
This is because non-parametric update can help in escaping fixed-points when they occur and the parametric update can help Alg.~\ref{alg:parallel-mma} in avoiding such fixed-points.
Combining both of these strategies further improves the results.
Secondly, we observe that performing message passing with GNN gives improvement over only using the MLP~$\Phi$.
Thirdly, we find using separate networks for early and late stage optimization gives better performance than using the same network for all stages. Lastly, using recurrent connections through an LSTM gives the best performance.

\paragraph{Limitations}
Easy problem classes, including small \textit{cell tracking}~\cite{haller2020primal} and easy Markov Random Field (MRF) inference~\cite{kappes2013comparative} do not benefit from learning, since \texttt{FastDOG} already solves the problem in few iterations. 
Some problem classes have sequential bottlenecks due to long subproblems, including MRFs for protein folding~\cite{jaimovich2006towards} and shape matching~\cite{windheuser2011geometrically,windheuser2011large}, which makes training difficult due to slow dual optimization. 

Although our method requires training for each problem class, the cost of training is manageable. Nonetheless devising a generalizable approach is an interesting research direction requiring at least: a large and diverse training set, powerful neural network and multi-GPU implementation. 



\section{Conclusion}
We have proposed an self-supervised learning approach for solving relaxations to combinatorial optimization problems by backpropagating through and learning parameters for the dual LP solver~\cite{abbas2021fastdog}.
We demonstrated its potential in obtaining close to optimal solutions much faster than with traditional methods.
Although our solvers require training as compared to conventional solvers, this overhead is negligible as compared to the human effort required for developing efficient specialized solvers (which are also often outperformed by our approach). 

Our work generalizes efficient approximate solver development: Instead of developing a specialized solver we propose to use a generically applicable one and train it to obtain fast and accurate optimization algorithm.
Going one step further and training a universal model that generalizes across different problem classes remains a challenge for future work.

\section*{Acknowledgements}
We thank all anonymous reviewers for their feedback especially reviewer 4 for the detailed discussion and pointing out related work. 
We also thank Paul Roetzer for suggestions regarding writing.
{
\small
\bibliography{references}
}

\newpage
\appendix
\onecolumn
\begin{center}
    \textbf{\Large Appendix}
\end{center}
\section{Proofs}
\subsection{Proof of Proposition~\ref{prop:dual-feasible-monotone}}

The proof is an adaptation of the corresponding proof for $\omega_{ij} = 0.5$ and $\alpha_{ij} = \frac{1}{\abs{\J_i}}$ given in~\cite{abbas2021fastdog}.
\begin{proof} 
\noindent\textbf{\\ Feasibility of iterates.}
We prove 
\begin{equation}
\sum_{j \in \J_i} \lambda^{j}_i + M_{ik} = c_i
\end{equation}
just after line~\ref{alg:after-forward-block-update} and~\ref{alg:after-backward-block-update} in Algorithm~\ref{alg:parallel-mma}.
We do an inductive proof over the number of iterates w.r.t iterations $t$.
\begin{description}
\item[$t = 0$:] 
\begin{itemize}
\item After~\ref{alg:after-forward-block-update}: Follows from $M = 0$ in line~\ref{alg:initialization-line}.
\item After~\ref{alg:after-backward-block-update}:
 Let $\lambda'$, $M'$, be the values that are used as input to line~\ref{alg:after-backward-block-update} and $\lambda$ and $M$ be the ones returned in line~\ref{alg:after-backward-block-update}.
It holds that
\begin{align}
\sum_{j \in \J_i} \left[\lambda_{ij} + M_{ij}\right]
=&
\sum_{j \in \J_i} \left[\lambda'_{ij} - M_{ij}(t) + \alpha_{ij} \sum_{k \in \J_i} \left(M'_{ik} \right)  + M_{ij} \right] \\
=& \sum_{j \in \J_i} \left[\lambda'_{ij} + M'_{ij}\right] \\
=& c_i\,.
\end{align}
by the proved inequality on $\lambda',M'$ and the assumption that $\sum_{k \in \J_i} \alpha_{ij} = 1$.
\end{itemize}
\item[$t > 0$:] Analoguously to the second point for $t = 0$.
\end{description}

\noindent\textbf{Non-decreasing Lower Bound.}
In order to prove that iterates have non-decreasing lower bound we will consider an equivalent lifted representation in which proving the non-decreasing lower bound will be easier.

\noindent\textbf{\\ Lifted Representation.}
Introduce $\lambda^{\beta}_{ij}$ for $\beta \in \{0,1\}$ and the subproblems
\begin{equation}
E(\lambda^{1}_{\bullet j}, \lambda^{0}_{\bullet j}) = \min_{x \in \X_j} x^\top \lambda^{1}_{\bullet j} + (1-x)^\top \lambda^{0}_{\bullet j}
\end{equation}
Then~\eqref{eq:dual-problem} is equivalent to
\begin{equation}
\label{eq:lifted-dual-problem}
\max_{\lambda^{1},\lambda^0} \sum_{j \in \J} E(\lambda^{1}_{\bullet j}, \lambda^{0}_{\bullet j}) \text{ s.t. } \sum_{j \in \J_i} \lambda^{\beta}_{ij} = \beta \cdot c_i
\end{equation}
We have the transformation from original to lifted $\lambda$
\begin{equation}
\lambda \mapsto (\lambda^1 \leftarrow \lambda, \lambda^0 \leftarrow \0)
\end{equation}
and from lifted to original $\lambda$ (except a constant term)
\begin{equation}
(\lambda^1, \lambda^0) \mapsto \lambda^1 - \lambda^0\,.
\end{equation}
It can be easily shown that the lower bounds are invariant under the above mappings and feasible $\lambda$ for~\eqref{eq:dual-problem} are mapped to feasible ones for~\eqref{eq:lifted-dual-problem} and vice versa. 

The update rule line~\ref{alg:lambda-update} in Algorithm~\ref{alg:parallel-mma} for the lifted representation can be written as
\begin{equation}
\label{eq:lifted-representation-lambda-update}
\lambda_{ij}^{\beta} \leftarrow \lambda_{ij}^{\beta} - \max((2\beta -1) M^{out}_{ij},0) + \alpha_{ij} \cdot \sum_{jk \in \J_i}\min((2\beta - 1) M_{ik}^{in},0)
\end{equation}
It can be easily shown that~\eqref{eq:lifted-representation-lambda-update} and line~\ref{alg:lambda-update} in Algorithm~\ref{alg:parallel-mma} are corresponding to each other under the transformation from lifted to original $\lambda$.

\noindent\textbf{\\ Continuation of Non-decreasing Lower Bound}
Define
\begin{equation}
\lambda_{ij}^{\prime \beta} = \lambda_{ij} - \omega_{ij} \cdot \max((2\beta - 1) (\min_{x \in \X_j: x_j = \beta} \la \lambda^{in}_{ij}, x \ra - \min_{x \in \X_j : x_i = 1-\beta} \la \lambda_{ij}^{in}, x \ra ), 0)\,.
\end{equation}
Then $E(\lambda^{\prime j,1}, \lambda^{\prime j,0}) = E(\lambda^{j,1}, \lambda^{j,0})$ 
are equal due to $\omega_{ij} \in [0,1]$.
Define next
\begin{equation}
\lambda_{ij}^{\prime \prime \beta} = \lambda_{ij}^{\prime } + \alpha_{ij} \sum_{k \in \J_i} \max((2\beta-1) M^{in}_{ik}, 0)\,.
\end{equation}
Then $E(\lambda^{\prime \prime j,1}, \lambda^{\prime \prime j,0}) \geq E(\lambda^{\prime j,1}, \lambda^{\prime j,0})$ since $\lambda'' \geq \lambda'$ elementwise. 
This proves the claim. 
\end{proof}

\subsection{Proof of Proposition~\ref{prop:half-iteration-grads}}
\begin{figure}
    \centering
    \begin{tikzpicture}[scale=1.0]
\tikzstyle{every node}=[font=\small, inner sep = 1pt]
\tikzstyle{mm_oper}=[chamfered rectangle, draw, color=black, fill=dmmacolor, thick, minimum height=0.75cm]
\tikzstyle{normal_oper}=[rectangle, draw, color=black, fill=gnncolor, thick, minimum height=0.75cm]
\tikzset{node distance = 1cm}
\tikzset{>= Stealth}

\node[style=mm_oper, align = center](m1) at (0, 0) {
    $\min\limits_{x \in \X_j: x_i = 1} \langle  \lambda_{\bullet j}^\old, x \rangle$
};

\node[style=mm_oper, align = center](m0) at (0, 1.5) {
    $\min\limits_{x \in \X_j: x_i = 0} \langle  \lambda_{\bullet j}^\old, x \rangle$
};

\node[draw=none, align=left](l_in) at (-2.75, 0.75) {$\lambda^\old$};
\node[draw=none, align=left](omega) at (-2.75, 2.25) {$\omega$};

\draw[in=180, out=0, ->, arrowcolor] (l_in) to node [draw=none, looseness=10, midway] (alpha1) {} (m0);

\draw[in=180, out=0, ->, arrowcolor] (l_in) to node [draw=none, looseness=10, midway] (alpha1) {} (m1);

\node[style=normal_oper, inner sep = 3pt](md) at (3.5, 0.75) {
    $\omega\cdot (m^1 - m^0)$
};

\draw[in=165, out=0, ->, arrowcolor] (m0) to node [draw=none, looseness=10, fill=white, midway] (alpha1) {$m^0$} (md);

\draw[in=195, out=0, ->, arrowcolor] (m1) to node [draw=none, looseness=10, fill=white, midway] (alpha1) {$m^1$} (md);


\draw[->, arrowcolor] (omega.0) -| (md.90);

\node[style=normal_oper, inner sep = 3pt](mm_avg) at (0,-1.7) {
    $\alpha_{ij}\sum_{k \in \J_i} M_{ik}^\old$
};

\node[draw=none, align=left](m_in) at (l_in |- mm_avg.166) {$M^\old$};
\node[draw=none, align=left, inner sep = 7pt](alpha) at (l_in |- mm_avg.194) {$\alpha$};

\draw[->, arrowcolor] (m_in.0) -- (mm_avg.166);
\draw[->, arrowcolor] (alpha.0) -- (mm_avg.194);

\node[circle, draw=black, fill=nonparamcolor, thick, below = 1cm of md](mm_minus) {
    \normalfont{$-$}
};

\node[circle, draw=black, fill=nonparamcolor, thick](mm_add) 
at (mm_avg -| mm_minus)
{
    \normalfont{$+$}
};

\draw[in=90, out=270, ->, arrowcolor] (md) to node [draw=none, looseness=10, midway] (m_out) {} (mm_minus);

\draw[in=90, out=270, ->, arrowcolor] (mm_minus) to node [draw=none, looseness=10, midway] (alpha1) {} (mm_add);

\draw[in=180, out=0, ->, arrowcolor] (mm_avg) to node [draw=none, looseness=10, midway] (alpha1) {} (mm_add);

\node[draw=none, right = 1.1cm of mm_add](l_out) {$\lambda^\new$};
\node[draw=none, right = 1.3cm of m_out](m_out_text) {$M^\new$};

\draw[->, arrowcolor] (m_out.center) -- (m_out_text.180);
\draw[->, arrowcolor] (mm_add.0) -- (l_out.180);

\draw[->, arrowcolor] (l_in.270) |- (mm_minus.180);

\end{tikzpicture}
    \caption{Computational graph of \texttt{BlockUpdate} in Alg.~\ref{alg:parallel-mma}}
    \label{fig:half-iteration-graph}
\end{figure}
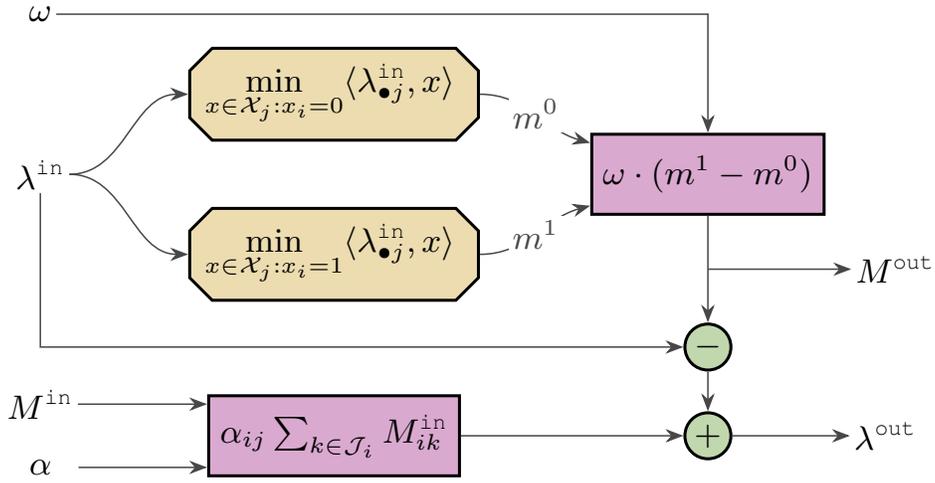
\begin{proof}
The computational graph of \texttt{BlockUpdate} in Alg.~\ref{alg:parallel-mma} is shown in Figure~\ref{fig:half-iteration-graph}. Assuming gradients 
$\nicefrac{\partial \mathcal{L}}{\partial M^\new}$ 
and $\nicefrac{\partial \mathcal{L}}{\partial \lambda^\new}$ are given. We first focus on lower part of Figure~\ref{fig:half-iteration-graph}. By applying chain rule gradient of $M^\old_{ij}\, \forall {ij} \in B$ is computed as
\begin{align}
\frac{\partial \mathcal{L}}{\partial M^\old_{ij}} 
= \sum_{p \in \I} \sum_{k \in \J_p} \frac{\partial \mathcal{L}}{\partial \lambda^\new_{pk}}\frac{\partial \lambda^\new_{pk}}{\partial M^\old_{ij}}
= \sum_{k \in \J_i} \frac{\partial \mathcal{L}}{\partial \lambda^\new_{ik}}
\frac{\partial \lambda^\new_{ik}}{\partial M^\old_{ij}}
= \sum_{k \in \J_i} \frac{\partial \mathcal{L}}{\partial \lambda^\new_{ik}}
\alpha_{ij}.
\label{eq:grad-def-mm}
\end{align}
Similarly gradient for $\alpha_{ij}\, \forall {ij} \in B$ is
\begin{align}
\frac{\partial \mathcal{L}}{\partial \alpha_{ij}} 
= \sum_{p \in \I} \sum_{k \in \J_p} \frac{\partial \mathcal{L}}{\partial \lambda^\new_{pk}}\frac{\partial \lambda^\new_{pk}}{\partial \alpha_{ij}}
= \frac{\partial \mathcal{L}}{\partial \lambda^\new_{ij}}\frac{\partial \lambda^\new_{ij}}{\partial \alpha_{ij}}
= \frac{\partial \mathcal{L}}{\partial \lambda^\new_{ij}}\sum_{k \in \J_i}M_{ik}^\old,
\label{eq:grad-alpha}
\end{align}
Since we allow running Alg.~\ref{alg:parallel-mma} for more than one iteration with same parameters ($\alpha$, $\omega$), the above gradient~\eqref{eq:grad-alpha} is accumulated to existing gradients of $\alpha$ to obtain the result given by Alg.~\ref{alg:parallel-mma-backprop}. 

For the upper part of Figure~\ref{fig:half-iteration-graph} we first backpropagate gradients of $\lambda^\new$ to $M^\new$ to account for subtraction ($-$) as
\begin{equation}
    \frac{\partial \mathcal{L}}{\partial M^\new} = \frac{\partial \mathcal{L}}{\partial M^\new} - \frac{\partial \mathcal{L}}{\partial \lambda^\new}.
\end{equation}

Then the gradient w.r.t.\ damping factors $\omega_{ij}\, \forall {ij} \in B$ is
\begin{equation}
    \frac{\partial \mathcal{L}}{\partial \omega_{ij}}
    = \frac{\partial \mathcal{L}}{\partial M^\new_{ij}}
    \frac{\partial M^\new_{ij}}{\partial \omega_{ij}}
    = \frac{\partial \mathcal{L}}{\partial M^\new_{ij}}
    \left(m^1_{ij} - m^0_{ij}\right) = \frac{\partial \mathcal{L}}{\partial M^\new_{ij}}
    \left(\frac{M^\new_{ij}}{\omega_{ij}}\right),
\end{equation}
which also needs to be accumulated to existing gradient as done for gradients of $\alpha$. 

Lastly to backpropagate gradients to $\lambda^\old$ we first calculate
\begin{subequations}
    \begin{align}
    \frac{\partial \mathcal{L}}{\partial m^0_{ij}} &= 
    \frac{\partial \mathcal{L}}{\partial M^\new_{ij}}
    \frac{\partial M^\new_{ij}}{\partial m^0_{ij}} = 
    -\frac{\partial \mathcal{L}}{\partial M^\new_{ij}}
    \omega_{ij}, \\ 
    \frac{\partial \mathcal{L}}{\partial m^1_{ij}} &= 
    \frac{\partial \mathcal{L}}{\partial M^\new_{ij}}
    \frac{\partial M^\new_{ij}}{\partial m^1_{ij}} = 
    \frac{\partial \mathcal{L}}{\partial M^\new_{ij}}
    \omega_{ij}.
    \end{align}
    \label{eq:mm-one-and-zero}
\end{subequations}
Then (sub-)gradient of min-marginals $m^0_{ij}, m^1_{ij}\, \forall {ij} \in B$ w.r.t.\ $\lambda^\old$ are
\begin{equation}
    \frac{\partial m^\beta_{ij}}{\partial \lambda} = \frac{\partial m^\beta_{ij}}{\partial \lambda_{\bullet j}} = 
    \argmin\limits_{x \in \X_j: x_{ij} = \beta} \la \lambda_{\bullet j}, x \ra, \quad \forall \beta \in \{0, 1\}.
    \label{eq:mm-jacobian}
\end{equation}
Using the above relations~\eqref{eq:mm-one-and-zero},~\eqref{eq:mm-jacobian} and applying chain rule we obtain
\begin{subequations}
    \begin{align}
    \frac{\partial \mathcal{L}}{\partial \lambda_{ij}^\old} &=
    \frac{\partial \mathcal{L}}{\partial \lambda_{ij}^\new} +
    \sum_{\beta \in \{0, 1\}}
    \sum_{p \in \I} \sum_{k \in \J_p} 
    \frac{\partial \mathcal{L}}{\partial m_{pk}^\beta}
    \frac{\partial m_{pk}^\beta}{\lambda_{ij}^\old} \\ 
    &=
    \frac{\partial \mathcal{L}}{\partial \lambda_{ij}^\new} + \sum_{\beta \in \{0, 1\}}
    \sum_{p \in \I_j} 
    \frac{\partial \mathcal{L}}{\partial m_{pj}^\beta}
    \frac{\partial m_{pj}^\beta}{\lambda_{ij}^\old}, \, \forall {ij} \in B.
    \end{align}
\end{subequations}
\end{proof}

\section{Efficient min-marginal computation and backpropagation}
Algorithms~\ref{alg:parallel-mma} and~\ref{alg:parallel-mma-backprop} in abstract terms require solving the subproblems each time a min-marginal value (or its gradient) is required. To make these procedures more efficient we represent each subproblem as binary decision diagrams (BDD) as done in~\cite{abbas2021fastdog}. We give a short overview below and refer to~\cite{abbas2021fastdog} for more details.
\paragraph{Binary decision diagrams (BDD).}
A BDD is a directed acyclic graph with arc set $A$ starting at a root node $r$ and ending at two nodes $\top$ and $\bot$.
For each variable $i$ the BDD contains one or more nodes in a set $\P_i$ where all $r\top$ paths pass through exactly one node in $\P_i$. 
All $r\top$ paths in the BDD correspond to feasible assignments of its corresponding subproblem.
Lagrange variables of the subproblem can be used as weights in BDD arcs allowing also to calculate cost of these $r\top$ paths.
This is done by creating two outgoing arcs for a node $v$ (except $\top$, $\bot$) in the BDD: a zero arc $vs^0(v)$ and a one arc $vs^1(v)$. If an $r\top$ path passes through zero arc $vs^0(v)$ it indicates that the corresponding variable has an assignment of $0$ and $1$ otherwise.

Therefore to compute the cost of assigning a $1$ to variable $i$ one needs to check all $r\top$ paths which make use of the one arcs from all nodes in $\P_i$. 
In~\cite{abbas2021fastdog} the authors compute min-marginals by maintaining shortest path distances. Each node $v$ in the BDD maintains the cost of shortest path from root node $r$ (denoted by $\shp(r,v)$) and cost of shortest path to $\top$ node. These path costs are updated in \normalfont{\texttt{BlockUpdate}} routine of Alg.~\ref{alg:parallel-mma}. Min-marginals $m^0, m^1$ for a variable $i$ in subproblem $j$ can be computed efficiently as
\begin{equation}
    \label{eq:min-marginal-via-shortest-path}
   m^{\beta}_{ij} = \min_{\substack{vs^{\beta}(v) \in A\\ v \in \P_i}} \left[\shp(r,v) + \beta\cdot\lambda_{ij} + \shp(s^{\beta}(v),\top)\right].
\end{equation}

Backpropagation through min-marginals $m^0$, $m^1$ can then be done by finding the $\argmin$ in~\eqref{eq:min-marginal-via-shortest-path} instead of the $\min$ operation. Afterwards the gradients can be passed to Lagrange variables $\lambda$ and shortest path costs $\shp(r, \cdot)$, $\shp(\cdot, \top)$ which minimize~\eqref{eq:min-marginal-via-shortest-path}.
Since shortest path costs are also computed by $\min$ operations (see Alg.\ 3, 4 in~\cite{abbas2021fastdog}), gradients of these path costs can subsequently be backpropagated to the Lagrange variables by the $\argmin$ operation. 

\paragraph{CUDA implementation:}
Although above-mentioned operations help towards an efficient implementation via scatter, gather operations available in Pytorch~\cite{pytorch} and Pytorch Geometric~\cite{pytorch_geometric}, GPU memory usage can still be high for large instances. This can be especially problematic for training since the GNN also needs a considerable amount of GPU memory. Therefore for further computational efficiency we implement both Algorithm~\ref{alg:parallel-mma} and its backpropagation in CUDA~\cite{cuda, Thrust} and expose via pybind~\cite{pybind11}. A comparison between our Pytorch implementation which relies on automatic differentation with our CUDA implementation is given in Table~\ref{tab:runtime_cuda}. We observe that for all datasets containing large instances (i.e., all datasets except \textit{Independent set}) GPU memory usage is drastically reduced through our CUDA implementation. Additionally, runtimes for both forward and backward pass are reduced. 
\begin{table*}[ht]
    \caption{Runtime and peak GPU memory usage statistics of one generalized min-marginal averaging iteration in Algorithm~\ref{alg:parallel-mma} and its backpropagation via Algorithm~\ref{alg:parallel-mma-backprop}. F.\ time: Runtime in milliseconds for forward pass i.e., one iteration of Alg.~\ref{alg:parallel-mma}, B.\ time: Runtime for its backpropagation, mem.\ : Maximum GPU memory in GB used during both forward and backward pass. The values are averaged over all training instances within each dataset.}
    \label{tab:runtime_cuda}
    \centering
    \resizebox{\textwidth}{!}{
    \begin{tabular}{l rrr rrr rrr rrr}
    \toprule
    \multirow{2}{*}{} & \multicolumn{3}{c}{\textit{Cell tracking}} & \multicolumn{3}{c}{\textit{Graph matching}} &  \multicolumn{3}{c}{\textit{Independent set}} & \multicolumn{3}{c}{\textit{QAPLib}} \\
    \cmidrule(lr){2-4} \cmidrule(lr){5-7} \cmidrule(lr){8-10} \cmidrule(lr){11-13}
     & F.\ time & B.\ time & mem.\  
     & F.\ time & B.\ time & mem.\ 
     & F.\ time & B.\ time & mem.\ 
     & F.\ time & B.\ time & mem.\
     \\
    \midrule
    PyTorch & 
    $305$ & $1039$ & $31$ & 
    $153$ & $344$ & $11$ & 
    $11$ & $16$ & $0.7$ & 
    $43.5$ & $70$ & $7.3$
    \\
    Our CUDA & 
    $16$ & $171$ & $3.4$ &
    $7$ & $68$ & $1.8$ & 
    $0.7$ & $5.7$ & $0.7$ & 
    $1.6$ & $16$ & $1.6$
    \\ 
    \bottomrule
    \end{tabular}
    }
\end{table*}

\section{Neural network details}
\subsection{Hyperparameters}
The hyperparameters used in experiments are reported in Table~\ref{tab:hypeparameters}. During training time we run the Alg.~\ref{alg:parallel-mma} for only a few iterations for computational efficiency since more iterations can make the backward pass much slower due to reverse mode autodiff. For test time we run Alg.~\ref{alg:parallel-mma} for more iterations since backward pass is not required. For \textit{QAPLib} dataset we need more training time than other datasets because training set is quite large and has more diversity as compared to other datasets.  To manage training we additionally use PyTorch lightning~\cite{pytorch_lightning}.
\begin{table*}[ht]
    \caption{
    Statistics of hyperparameters of our approach for each dataset. 
    $T$: Num. of iterations of Alg.~\ref{alg:parallel-mma} in each optimization round; $R$: max.\ number of training rounds; \# itr.\ train: Num. of training iterations.}
    \label{tab:hypeparameters}
    \centering
    \begin{tabular}{l cccc ccc}
    \toprule
        \multirow{2}{*}{Dataset} & 
        \multicolumn{2}{c}{$T$} &
        \multirow{2}{*}{$R$} &
        \multirow{2}{*}{\parbox{1.0cm}{\centering batch size}} &
        \multirow{2}{*}{\parbox{1.0cm}{\centering learn.\ rate}} &
        \multirow{2}{*}{\parbox{1.2cm}{\centering \# itr.\ train}} &
        \multirow{2}{*}{\parbox{1.4cm}{\centering train time [hrs]}} \\
        \cmidrule(lr){2-3} 
         & train & test & & & \\
    \midrule
        \textit{CT} & $1$ & $100$ & $400$ & $1$ & 1e-3 & $500$ & $14$ \\
        \textit{GM} & $20$ & $200$ & $20$ & $2$ & 1e-3 & $400$ & $4$ \\
        \textit{IS} & $20$ & $50$ & $20$ & $8$ & 1e-3 & $2500$ & $10$ \\
        \textit{QAPLib} & $5$ & $20$ & $500$ & $4$ &  1e-3 & $1600$ & $48$ \\
    \bottomrule
    \end{tabular}

\end{table*}

\subsection{Hand-crafted features}
The features used as input to the neural networks at every optimization round are provided in Table~\ref{tab:gnn_features}. The motivation of using these features is as follows
\begin{enumerate}
    \item 
    For GNN to have complete ILP description we encode primal objective vector, constraint matrix, right-hand side vector, and constraint type. For more expressiveness we encode node degrees separately. 
    \item 
    We provide GNN all of the information which FastDOG~\cite{abbas2021fastdog} uses to perform dual updates. For this purpose we encode Lagrange variables $\lambda$ and deferred min-marginal differences $M$. 
    \item Features which can be computed easily and can potentially help the GNN. For example the optimal solution of each subproblem which actually corresponds to super-gradient of the dual objective~\eqref{eq:dual-problem} is given. In addition we provide gradients of smoothed dual objective~\eqref{eq:dual-problem}. To this end we replace each subproblem~\eqref{eq:dual_subproblem} with its smoothed variant as
    \begin{equation}
    E_{\alpha}^j(\lambda_{\bullet j}) = \alpha\cdot \log\left(\sum_{x \in \X_j}\exp\left( \frac{\la  \lambda_{\bullet j}, x \ra}{\alpha} \right)\right),
    \label{eq:smooth_dual_subproblem}
    \end{equation}
    with varying values of smoothing factor $\alpha$. For more details we refer to Sec.\ A.5 of~\cite{lange2021efficient}.
    \item Moving average of previously computed features. This can help the GNN to be informed about the goodness of past updates, thus allowing for (implicit) change of step-sizes.  
\end{enumerate}
\begin{table}[h]
    \caption{Features used for learning. Exponentially averaged features are computed with a smoothing factor of $0.9$. Features corresponding to the ILP remain fixed (i.e.\ node degrees, constraint type, $c$, $A$, $b$) whereas the remaining features are updated after every optimization round. }
    \label{tab:gnn_features}
    \centering
    \begin{tabular}{lll}
    \toprule
    Types & Feature description \\
    \toprule
    \multirow{2}{*}{Primal variables $f_\I$} & Normalized cost vector $c / \lVert c \rVert_\infty$ \\
    & Node degree ($\abs{\J_i}\, \forall i \in \I$) \\
    \midrule
    \multirow{6}{*}{Subproblems $f_\J$} 
    & Node degree ($\abs{\I_j}\, \forall j \in \J$) \\
    & RHS vector $b$ in constraints $Ax \leq b$ \\
    & Indicator for constraint type ($\leq$ or $=$) \\
    & Current objective value per subproblem $[E^1(\lambda_{\bullet 1}), \dotsc, E^m(\lambda_{\bullet j})]$ \\
    & Exp.\ moving avg.\ of first, second order change in obj.\ value \\
    & Change in objective value due to last non-parametric update~\eqref{eq:non-parametric-update} \\ 
    \midrule
    \multirow{5}{*}{Dual variables $f_\mathcal{E}$}
    & Current optimal assignment of each subproblem (i.e., subgradient of dual objective~\eqref{eq:dual-problem}) \\
    & Gradients of the smoothed dual objective using~\eqref{eq:smooth_dual_subproblem} for all $\alpha$ in $\{1.0, 10.0, 100.0\}$.\\
    & Exp.\ moving avg.\ of optimal assignment \\
    & Coefficients of constraint matrix $A$ \\
    & Current (normalized) Lagrange variables ${\lambda} / {\lVert \lambda + M + \epsilon \rVert}$ \\
    & Current (normalized) deferred min-marginal differences ${M} / {\lVert \lambda + M + \epsilon \rVert}$ \\
    \bottomrule
    \end{tabular}
\end{table}

\section{Detailed results}
\subsection{Generalization study}
\label{subsec:generalization}
To evaluate the generalization power we test our approach on different types of instances within a problem class. To this end we take the \textit{QAPLib} dataset which naturally contains instances of different types of problems e.g., keyboard layout design, facility location, circuit design etc. We train our solver on each sub-category and test on all categories. Note that for training we take our original training split and subdivide into different sub-categories. The results are available in Table~\ref{tab:qaplib_generalization}.

We observe only a few cases where our approach generalizes to instances of different types e.g. training on \textit{tai$^*$} generalizes to \textit{lipa} but not the other way around. We hypothesize that it is due to each dataset having limited diversity causing overfitting.

\setlength{\fboxsep}{2mm} 
\setlength{\tabcolsep}{0pt}
\begin{table}[h]
    \caption{
    Results of training and testing on different problem types within the \textit{QAPLib} dataset. Values in the table depict relative dual gap (lower values are better). We divide our original training and testing sets into smaller splits where each split contains instance of only one problem type. Each row then indicates a training run on one training split and the columns indicate results on different test instances. The instances used in training splits are
    \textit{bur$^*$}: \textit{bur26a-d},
    \textit{lipa$^*$}: \textit{lipa$\{$20a-b, 30a-b$\}$},
    \textit{nug$^*$}: \textit{nug$\{$12, 14, 15, 16a-b, 17, 18, 20, 21, 22, 24, 25$\}$},
    \textit{tai$^*$}: \textit{tai$\{$10a-b, 12a-b, 15a-b, 17a, 20a-b, 25a-b, 30a-b$\}$},
    \textit{chr$^*$}: \textit{chr$\{$12a-c, 15a-c, 18a-b, 20a-c, 22a$\}$};
    \textit{full}: Our full training split of \textit{QAPLib} as defined in Sec~\ref{subsec:datasets} (does not contain any test instance of this experiment).
    }
    \label{tab:qaplib_generalization}
    \begin{center}
        \begin{tabular}{l | *{13}{R}}
        \diagbox[height=10mm, width=11mm]{Train}{Test}
        & \rotatebox{45}{\textit{bur26e-h}} & \rotatebox{45}{\textit{lipa40a-b}} & \rotatebox{45}{\textit{nug{28,30}}} & \rotatebox{45}{\textit{tai35a-b}} & \rotatebox{45}{\textit{chr22b}} & \rotatebox{45}{\textit{had20}} & \rotatebox{45}{\textit{kra32}} & \rotatebox{45}{\textit{rou20}} & \rotatebox{45}{\textit{scr20}} & \rotatebox{45}{\textit{ste36c}} & \rotatebox{45}{\textit{tho40}} \EndTableHeader\\
          \hline
          \textit{bur$^*$} \,\, & 0.0073 & 0.3220 & 0.3394 & 0.6180 & 0.0244 & 0.2129 & 0.0601 & 0.3760 & 0.0116 & 0.1313 & 0.4951 \\
          
          \textit{lipa$^*$} \,\, & 0.3371 & 0.0024 & 0.0218 & 0.2647 & 0.0530 & 0.0003 & 0.0099 & 0.0024 & 0.0493 & 0.1198 & 0.0629  \\
          
          nug$^*$ \,\, & 0.0056 & 0.0839 & 0.1903 & 0.3390 & 0.0161 & 0.0325 & 0.0348 & 0.1185 & 0.0111 & 0.0778 & 0.1913 \\
          
          \textit{tai$^*$} \,\, & 0.0697 & 0.0067 & 0.0191 & 0.0009 & 0.0803 & 0.0001 & 0.0306 & 0.0000 & 0.0217 & 0.2430 & 0.0964 \\

          \textit{chr$^*$} \,\, & 0.0914 & 0.0185 & 0.0349 & 0.3050 & 0.0212 & 0.0323 & 0.0354 & 0.0472 & 0.0094 & 0.1023 & 0.0584 \\

          \textit{full} \,\, & 0.0026 & 0.0056 & 0.0009 & 0.0096 & 0.0358 & 0.0000 & 0.0126 & 0.0000 & 0.0022 & 0.0345 & 0.0491 \\
        \end{tabular}
    \end{center}
\end{table}
\setlength{\tabcolsep}{4pt}
\subsection{Ablation study}
Extended results of ablation study in Section~\ref{subsec:results} on the datasets of \textit{Independent set} and a smaller split (for efficiency reasons) of \textit{QAPLib} are given in Table~\ref{tab:ablation_is_qaplib}. As before we observe that learning both parametric updates by our Alg.~\ref{alg:parallel-mma} and non-parametric updates~\eqref{eq:non-parametric-update} are essential for good performance. Moreover learning even one of these components allows us to surpass the hand-designed algorithm \texttt{FastDOG}~\cite{abbas2021fastdog}.

\setlength{\tabcolsep}{1pt}
\begin{table*}[ht]
\caption{Ablation study results on \textit{Independent set} and \textit{QAPLib} datasets.
\texttt{FastDOG}: Non-learned baseline from~\cite{abbas2021fastdog};
only non-parametric: predict only the non-parametric update~\eqref{eq:non-parametric-update}, only parametric: predict only the parameters of Algorithm~\ref{alg:parallel-mma} without non-parametric update.}
\label{tab:ablation_is_qaplib}
\centering
    \begin{tabular}{c c c c c c c c c} 
        \toprule  
        & \multicolumn{4}{c}{\textit{Independent set}} & \multicolumn{4}{c}{\textit{QAPLib}} \\
        \cmidrule(lr){2-5} \cmidrule(lr){6-9} 
         & \multirow{2}{*}{\parbox{1.5cm}{\centering w/o learn. \\ (\texttt{FastDOG})}} &
        \multirow{2}{*}{\parbox{3cm}{\centering only \\ non-parametric}} & \multirow{2}{*}{\parbox{2cm}{\centering only parametric}} &
        \multirow{2}{*}{\texttt{DOGE-M}}
         & \multirow{2}{*}{\parbox{2.5cm}{\centering w/o learn. \\ (\texttt{FastDOG})}} &
        \multirow{2}{*}{\parbox{3cm}{\centering only \\ non-parametric}} & \multirow{2}{*}{\parbox{2cm}{\centering only parametric}} &
        \multirow{2}{*}{\texttt{DOGE-M}}\\ 
        & & & & & & & & \\
        \midrule
        $g_I$ $(\downarrow)$ & $14$ 
        & $3.12$ & $1.76$ 
        & $\textbf{0.2}$ & $1310$ 
         & $641$ & $564$ 
         & $\textbf{301}$\\
        \midrule

        $E$ $(\uparrow) $ & $-24913$ & $-24913$ & $-24472$ & $-\textbf{24459}$ 
        & $3.9e6$ & $4.3e6$ & $3.6e6$ & 
        $\textbf{6.8e6}$ \\
        \bottomrule
    \end{tabular}
\end{table*}
\setlength{\tabcolsep}{4pt}


\subsection{Results on smaller instances of \textit{QAPLib}}
\begin{figure}[h]
    \centering
\includegraphics[width=0.7\textwidth]{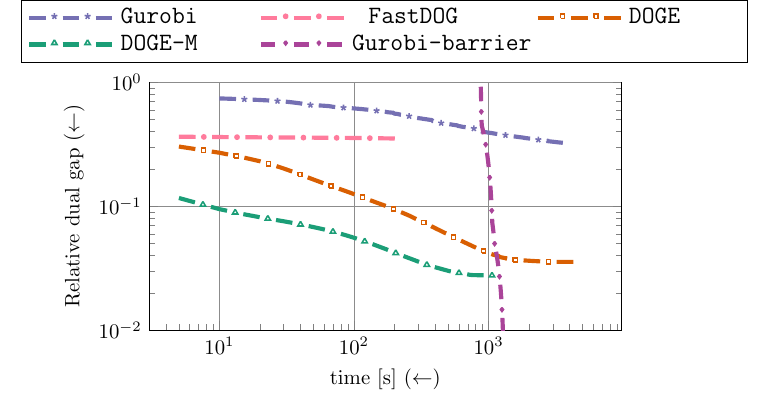}
    \caption{Convergence plots of smaller test instances of \textit{QAPLib} ($\leq$ 40 nodes).}
    \label{fig:qaplib_small_until_40}
\end{figure}

In Figure~\ref{fig:qaplib_small_until_40} we provide additional convergence plot calculated only on smaller instances of \textit{QAPLib} dataset. These instances contain on average $1.6$ million dual variables (instead of the overall test split with $11$ million). We observe that on relatively smaller instances our solvers \normalfont{\texttt{DOGE}}, \normalfont{\texttt{DOGE-M}} are surpassed by the barrier method but not by the dual simplex method of~\cite{gurobi}. However, on larger instances the barrier method could not perform any iteration within $1$ hour timelimit. 

\clearpage
\subsection{\textit{Cell tracking}}
\insertTable{figures/tables/ct_logs_merged.csv}{Detailed results on \textit{Cell tracking} dataset. Until termination criteria contain results where we stop our solvers early w.r.t.\ relative improvement. These results are averaged and reported in Table~\ref{tab:combined-results}. Best until max.\ itr.: We run our solver for at most 50000 iterations and report best results (so $R$ = 500, $T$ = 100).}
{tab:ct-instance-results}

\subsection{\textit{Graph matching}}
\insertTable{figures/tables/worm_logs_merged.csv}{Detailed results on \textit{Graph matching} dataset. Until termination criteria contain results where we stop our solvers early w.r.t.\ relative improvement. These results are averaged and reported in Table~\ref{tab:combined-results}. Best until max.\ itr.: We run our solver for at most 10000 iterations and report best results (so $R$ = 50, $T$ = 200).}
{tab:gm-instance-results}

\subsection{\textit{QAPLib}}
\insertTable{figures/tables/qaplib_all_instances_logs_merged.csv}{Detailed results on \textit{QAPLib} dataset. Until termination criteria contain results where we stop our solvers early w.r.t.\ relative improvement. These results are averaged and reported in Table~\ref{tab:combined-results}. Best until max.\ itr.: We run our solver for at most 100000 iterations and report best results (so $R$ = 5000, $T$ = 20). *: Gurobi did not converge within 1 hour timelimit. }{tab:qaplib-instance-results}


\end{document}